\documentclass[preprint,aps,pre,letterpaper,onecolumn,superscriptaddress]{revtex4-2} 

\usepackage{algorithm}
\usepackage{algpseudocode}
\usepackage[margin=1in]{geometry}
\usepackage{graphicx}
\usepackage{amssymb}
\usepackage{amsmath}
\usepackage{mathrsfs}
\usepackage{placeins}
\usepackage{caption3}
\usepackage{wrapfig}
\usepackage{caption} 
\usepackage{subcaption}
\usepackage{xcolor}
\usepackage{contour}
\usepackage{color}
\usepackage{outlines}
\usepackage[bookmarksnumbered=true,breaklinks=true,colorlinks,citecolor=blue,linkcolor=blue]{hyperref}

\usepackage[normalem]{ulem}

\newlength{\figwidth}
\newlength{\SCwidth}
\def\XXint#1#2#3{{\setbox0=\hbox{$#1{#2#3}{\int}$}
		\vcenter{\hbox{$#2#3$}}\kern-.5\wd0}}

\definecolor{ao(english)}{rgb}{0.0, 0.5, 0.0}
\newcommand{\ALrevise}[1]{\textcolor{black}{#1}}
\begin{document}


\title{Stabilized Neural Ordinary Differential Equations for Long-Time Forecasting of Dynamical Systems}


\author{Alec J. Linot}
\email{linot@wisc.edu}
\affiliation{Department of Chemical and Biological Engineering, University of Wisconsin-Madison, Madison WI 53706, USA}

\author{Joshua W. Burby}
\affiliation{Los Alamos National Laboratory, Los Alamos, NM 87545, United States of America}

\author{Qi Tang}
\affiliation{Los Alamos National Laboratory, Los Alamos, NM 87545, United States of America}

\author{Prasanna Balaprakash}
\affiliation{Mathematics and Computer Science Division, Argonne National Laboratory, Lemont, IL 60439, USA}

\author{Michael D. Graham}
\affiliation{Department of Chemical and Biological Engineering, University of Wisconsin-Madison, Madison WI 53706, USA}

\author{Romit Maulik}
\affiliation{Mathematics and Computer Science Division, Argonne National Laboratory, Lemont, IL 60439, USA}


\date{\today}

\begin{abstract}

\ALrevise{In data-driven modeling of spatiotemporal phenomena} careful consideration often needs to be made in capturing the dynamics of the high wavenumbers. This problem becomes especially challenging when the system of interest exhibits shocks or chaotic dynamics. We present a data-driven modeling method that accurately captures shocks and chaotic dynamics by proposing a novel architecture,~\emph{stabilized neural ordinary differential equation} (ODE). In our proposed architecture, 
\ALrevise{we learn the right-hand-side (RHS) of an ODE by adding the outputs of two NN together where one learns a linear term and the other a nonlinear term. Specifically, we implement this by training a sparse linear convolutional NN to learn the linear term and a dense fully-connected nonlinear NN to learn the nonlinear term.}
This is in contrast with the standard neural ODE which involves training only a single NN for learning the RHS.  We apply this setup to the viscous Burgers equation, which exhibits shocked behavior, and show better short-time tracking and prediction of the energy spectrum at high wavenumbers than a standard neural ODE. We also find that the \ALrevise{stabilized neural ODE models are much more robust to noisy initial conditions than the standard neural ODE approach.}
\ALrevise{We also} apply this method to chaotic trajectories of the Kuramoto-Sivashinsky equation. In this case, \ALrevise{stabilized neural ODEs keep long-time trajectories on the attractor, and are highly robust to noisy initial conditions, while standard neural ODEs fail at achieving either of these results.}
\ALrevise{We conclude by demonstrating how stabilizing neural ODEs provide a natural extension for use in reduced-order modeling by projecting the dynamics onto the eigenvectors of the learned linear term.}

\end{abstract}


\maketitle


\section{Introduction}



With the ever increasing amount of data available from experiments or high-resolution simulations it has become common to build models directly and strictly from data. This tactic is powerful in that no {\it a priori} knowledge of the system is needed and it forces the data to guide the solution. 
Unfortunately, models built \ALrevise{without any knowledge} of the system often lack robustness to perturbations in the data and do a poor job of generalizing \ALrevise{outside of the training data.}
\ALrevise{In particular, we are interested in the problem of modeling dynamical systems from time-series data. For example, this could be predicting the future location of a pendulum given angle and velocity data, predicting the weather, or predicting chemical concentrations in a reactor.}
\ALrevise{Time-series predictions are especially susceptible to issues of robustness because states are repeatedly put through the model and error that pushes states away from the training data become worse as the state moves further and further away from the training data.}


Here we seek to overcome the issue of robustness, in many data-driven models, by using the fact that often in dissipative partial differential equations (PDEs), stability comes from a linear operator that dissipates energy \cite{Constantin1989}.
\ALrevise{In particular, we are concerned with the \emph{nonlinear} stability of attractors. 
By nonlinear stability we mean finite perturbations to states on the attractor are pushed back to the attractor. This type of stability can be show using the energy method \cite{Straughan1992}, with Lyapunov functions \cite{Khalil2002}, or for dynamical systems which possess an inertial manifold \cite{Constantin1989}. The energy method in particular highlights the importance of the linear term in stabilizing many dissipative PDEs.} \ALrevise{For example, when computing the energy balance for the Viscous Burgers Equation (VBE) and the Navier-Stokes Equation dissipation comes from the diffusive term \cite{Pope2000} and for the Kuramoto-Sivashinky equation (KSE) hyperdiffusivity dissipates energy \cite{Cvitanovic2012}. Below we make this notion of stability more precise.}



\ALrevise{In what follows, we briefly review some data-driven modeling approaches and then outline our stabilizing neural ODE approach. In this approach we train a linear NN and a nonlinear NN simultaneously to approximate the right hand side of an ODE, which we find stabilizes the system to perturbations. None of the existing approaches described explicitly learn a linear term in an ODE from snapshots of data.}


When \ALrevise{ modeling time-series data of} autonomous systems \ALrevise{(i.e. the right hand side does not explicitly depend on time)} 
two approaches are to either represent the dynamics as discrete time flows
\begin{equation}\label{eq:discrete}
	u(t+\tau)=F(u(t)),
\end{equation}
or as an ODE
\begin{equation}\label{eq:ODE}
	\dfrac{d u}{d t}=f(u),
\end{equation} 
where $u$ is the state of the system at a given time.
\ALrevise{Often, snapshots of data, in the form $\{u(t_1),\ldots,u(t_N)\}$, will be available for training a model.}
\ALrevise{This makes discrete time flows a natural first approach because that is the form in which the data is available (i.e. the data is in the precise form that a discrete time flows is capable of predicting).} 

When the dynamics are simple, a common approach to estimating $F$ is through a linear map \ALrevise{(i.e. $u(t+\tau)=Gu(t)$)} using dynamic mode decomposition (DMD) \cite{DMDBook}. 
DMD works as a predictive tool for systems that decay to a fixed point or exhibit quasiperiodic dynamics with discrete frequencies. \ALrevise{When applying DMD, or extended DMD (DMD with nonlinear observables), to nonlinear systems challenges arise in the form of closure of models and robustness to noise \cite{Wu2021}.} \ALrevise{Lusch et al. \cite{Lusch2018} showed one way to extend these ideas for a few simple nonlinear systems is to use an autoencoder to learn a change of basis under which the dynamics are linear. Moreover, even with more complicated nonlinear systems there always exists a linear \emph{Koopman operator} \cite{Budisic2012}. The Koopman operator describes the evolution of arbitrary observables, but is infinite-dimensional. Due to these issues, we do not seek a strictly linear time evolution method. We instead consider methods that include nonlinearity.}



Two classes of machine learning approaches that have seen success in modeling nonlinear systems include reservoir networks \cite{Lukosevicius2009,Pathak2018a} and recurrent neural networks (RNN) \cite{Hochreiter1997,Vlachas2018,Vlachas2019}. Both of these methods work by finding the optimal parameters to some function that maps $u(t)$ forward to some $\tilde{u}(t+\tau)$ by using a history of passed states. The parameters are then trained by minimizing the difference between the prediction and the true state $\left<||u(t+\tau)-\tilde{u}(t+\tau)||^2\right>$. Both methods can predict multiple Lyapunov times for the chaotic dynamics of the KSE
\cite{Vlachas2019}. \ALrevise{The Lyapunov time is the relevant time scale for chaotic systems and will be discussed in more detail below.} Some drawbacks of these methods are they typically result in an increase in the dimension of the problem, they typically require evenly spaced data, and the start up typically needs multiple known states \ALrevise{because these methods are non-Markovian (e.g. predictions of the next state require multiple past states)}. 


Instead of learning a discrete-time representation it is also possible to learn an ODE. In general, this is less straightforward because the data available tends to be the state at different points in time. However, if the time derivative is available Gonzalez Garcia et al.\ \cite{Gonzalez-Garcia1998} showed that a NN with states and derivative inputs could learn the ODE.
\ALrevise{A similar, but more interpretable approach is ``Sparse Identification of Nonlinear Dynamics" (SINDy)\cite{Brunton2016}. This approach involves selecting a dictionary of candidate functions for Eq.\ \ref{eq:ODE}, and using sparse regression to identify the dominant terms which most accurately reconstruct the time derivative.} Alternatively, the time derivative can be approximated \ALrevise{with a multistep time-integration scheme} and a NN can be trained off of this approximation \cite{Raissi2018}. \ALrevise{A drawback of all these approaches is the data needs to be close in time to get good approximations of the derivative, if the time derivative is not known.}

The data used can be spaced further in time by using the ``neural ODE" approach developed by Chen et al.\ \cite{Chen2019}. In this approach a trajectory is evolved forward using a NN for $f$ with some numerical time integrator. Then the NN is trained to minimize the difference between this prediction and the true state. In Section \ref{sec:Framework} we outline this approach in more detail. The advantages this approach has over the other methods include data can be spaced further in time, trajectories can be evolved to arbitrary points in time, and adaptive time stepping can be used.

Neural ODEs have been successfully used for short-time prediction of Burgers equation \cite{MAULIK2020}, for the evolution of dissipation in decaying isotropic turbulence \cite{portwood2019turbulence}, and for flow around a cylinder \cite{rojas2021reducedorder}. Additionally, in \cite{Linot2021} we showed that the long-time dynamics of the KSE could be captured using neural ODEs. However, this only works when the dimension of the problem is reduced. Without any dimension reduction high wavenumbers are amplified resulting in trajectories leaving the attractor. In this work we seek to overcome this problem of high-wavenumbers amplifying not by dimension reduction, but by explicitly adding a linear term into the neural ODE approach. \ALrevise{The properties of this linear term are described below.} 

\ALrevise{A similar idea to learn a linear term of an ODE is ``Linear and Nonlinear Disambiguation Optimization" (LANDO) \cite{Baddoo2021}. In this approach Eq.\ \ref{eq:ODE} is approximated with a dictionary-based kernel model. Then, the linear term on the RHS of the ODE is approximated by calculating the Jacobian of $f$ around the zero base state. There are two clear practical differences between that approach and the one we present. First, LANDO requires either time derivative data, or data closely spaced in time, whereas neural ODEs can use data widely spaced in time. Second, the Jacobian around the base state is only accurate if the function approximation is accurate around the base state. In our case, the base state is the zero solution, and the data we use for training never approaches the base state, so the Jacobian around this state may be poorly estimated.}





In what follows, we show that by adding an explicit linear term into the neural ODE we can improve long-time predictive capabilities, we can improve robustness to noise, and we can create a reduced order model from the basis given by the linear term. Section \ref{sec:Framework} describes the framework we take for constructing and training our models. Then, in Section \ref{sec:Results} we present the performance of the different models on the VBE (Section \ref{sec:VBE}) and on the KSE (Section \ref{sec:KSE}). The VBE is presented to show how the model can handle shock behavior and the KSE is presented to show how the model handles prediction of chaotic dynamics on an attractor. A key result in both situations is \ALrevise{for standard neural ODEs, that lack an explicit linear term,} 
high wavenumbers grow in time, but by learning a linear term this erroneous behavior is avoidable and the models become much more robust to noise. Finally, we \ALrevise{conclude} in Section \ref{sec:Conclusion}.

\section{Stabilized Neural ODE Framework} \label{sec:Framework}


We first introduce our proposed ~\emph{stablized neural ODE} framework, which builds a dissipative linear mapping into the conventional neural ODE framework. The approach is motivated by \ALrevise{the stabilizing effect} the linear term has in many dynamical systems. We further discuss a natural and interesting extension of the proposed approach into reduced order modeling (ROM).

For our analysis we consider snapshots of data $\left\{u(t_1), u(t_2), \ldots, u(t_N)\right\}$ that come from solutions to an autonomous 
ODE with the form
\begin{equation}\label{eq:ODEFull}
	\dfrac{du}{dt}=f(u)=Au+F(u).
\end{equation}
Here $u(t_i)\in \mathbb{R}^d$ is the state, $t\in \mathbb{R}$ is time, $A\in \mathbb{R}^{d\times d}$ is a matrix, and $f:\mathbb{R}^d \rightarrow \mathbb{R}^d$ and $F:\mathbb{R}^d \rightarrow \mathbb{R}^d$ are nonlinear functions.
\ALrevise{In particular we investigate systems where $A$ is symmetric and $F$ contains a quadratic nonlinearity, but the ideas should hold for any system with a dissipative linear term.}
\ALrevise{For the two equations we consider, the VBE and the KSE in one spatial dimension, we know, not only that the long-time solutions lie on attractors, but that the long-time dynamics also collapse onto inertial manifolds \cite{Vukadinovic2011,Foias1988a}. An inertial manifold is a finite-dimensional invariant attractive manifold that attracts trajectories exponentially fast \cite{Foias1988b}.}


With this data, we seek to train a NN to approximate the RHS of this ODE using the neural ODE framework. Typically only $f$ is represented by a NN, but here we show that a better approach is training two NNs \emph{simultaneously}, a linear NN \ALrevise{-- in our case this will be a convolutional NN (CNN) --} 
for $A$ and a nonlinear NN for $F$. When we approximate one of these terms with a NN, or make a prediction with these approximations, we will denote the approximation with \ALrevise{a tilde} (e.g. $\tilde{f}$). 

\ALrevise{For our trials we constrain $A$ to be symmetric for the VBE, and apply no constraints for the KSE. We chose to limit the constraints applied to the neural ODE because the underlying systems possesses nonlinear stability in the form of an inertial manifold. We took this approach to show that the model can learn this property from the data when the neural ODE is provided a linear term. An alternative future direction is explicitly enforcing the existence of an inertial manifold. This could be done by enforcing a spectral gap condition, which guarantees the existence of an inertial manifold \ALrevise{\cite{Zelik2014}}. The spectral gap condition is}

\begin{equation}
    \lambda_{N+1}-\lambda_{N}>2 L
\end{equation}
\ALrevise{
where $\lambda_i$ is the $i^{\text{th}}$ eigenvalue of $A$, sorted in increasing order, and $L$ is the Lipschitz constant of $F$.
}

To train these NNs, we numerically integrate one of the ODEs forward in time to estimate the state $\tilde{u}(t_i+\tau)$
\begin{equation}\label{eq:ODENet_Int}
	\tilde{u}(t_i+\tau)=u(t_i)+\int_{t_i}^{t_i+\tau}\tilde{h}(u(t);\theta) dt.
\end{equation}
\ALrevise{In what follows we consider three variations on $\tilde{h}$: 1) a nonlinear NN, $\tilde{h}(u)=\tilde{f}(u;\theta_1)$, 2) a nonlinear NN with a fixed linear term, $\tilde{h}(u)= Au+\tilde{F}(u;\theta_2)$, and 3) a nonlinear NN with a linear CNN, $\tilde{h}(u)=\tilde{A}(\theta_3)u+\tilde{F}(u;\theta_2)$. Variations 2 and 3 are what we refer to as stabilized neural ODEs. 
\ALrevise{In variation 2 we use the true linear term of the underlying system, however the framework is agnostic to this selection. For example, simply replacing $A$ with a scalar damping term may be useful for model stability.}
In all of these cases, $\theta_i$ refers to the parameters that are trained.}
The integration of $\tilde{h}$ is then used to calculate the loss, which is the difference between the predicted state, $\tilde{u}(t_i+\tau)$, and the known state, $u(t_i+\tau)$
\begin{equation}\label{eq:ODENet}
	J=\left<||u(t_i+\tau)-\tilde{u}(t_i+\tau)||_1\right>.
\end{equation}
We use the $L_1$-norm, but other norms are also applicable. 

For learning the parameters of these NNs, the gradient of this loss with respect to all the parameters, $\theta$, must be calculated. This gradient can either be calculated by backpropagating through the solver with automatic differentiation or by solving an adjoint problem backwards in time \cite{Chen2019}. When backpropagating all of the data must be stored at each time-step, which becomes memory intensive when the \ALrevise{prediction horizon becomes large.} For our trials we do not consider \ALrevise{training} data spaced apart far enough in time where this becomes a problem, and empirically we found using the adjoint method took longer, so we use backpropagation for training. 


In addition to stabilizing the dynamics, which we show in the following section, the addition of a linear term leads to an interesting and novel ROM framework, i.e., the stablized neural ODE framework implies a natural basis for ROM \ALrevise{based on the eigenvectors of the linear term.}
\ALrevise{This ``flips" the traditional ROM approach in that we first discovery dynamics and then use this dynamical system with classical methods for ROM. In particular, we apply standard Galerkin and nonlinear Galerkin approaches for ROM. These methods require that the ODE in Eq.\ \ref{eq:ODEFull} is known, but because we learn this equation from data we can directly apply these methods without knowing the equation a priori.}

\ALrevise{Specifically, we perform an eigendecomposition on the linear term, which we learn from a CNN, to find} $\tilde{A}\mathrm{v}_i=\lambda_i \mathrm{v}_i$ and then \ALrevise{we} project \ALrevise{the model approximation of} Eq.\ \ref{eq:ODEFull} onto the leading eigenvectors, giving the~\emph{resolved} dynamics
\begin{equation}\label{eq:resolved}
	\frac{\mathrm{d} p}{\mathrm{~d} t} = \tilde{A} p + P \tilde{F}(p+q),
\end{equation}
and the trailing eigenvectors, giving the~\emph{unresolved} dynamics
\begin{equation}\label{eq:unresolved}
	\frac{\mathrm{d} q}{\mathrm{~d} t} = \tilde{A} q + Q \tilde{F}(p+q),
\end{equation}
Here $p=Pu$ is the projection onto the leading eigenvectors (resolved dynamics) and $q=Qu$ is the projection onto the trailing eigenvectors (unresolved dynamics). 
\ALrevise{For the systems we consider, $\tilde{A}$ is symmetric, which results in real eigenvalues and orthogonal eigenvectors. This means the projections can be written $P=V_pV_p^T$ and $Q=V_qV_q^T$, where $V_p=[\mathrm{v}_0,\dots,\mathrm{v}_p]$ and $V_q=[\mathrm{v}_{p+1},\dots,\mathrm{v}_d]$.} The classic way to sort the eigenvectors is in the order of decreasing real part of the eigenvalues \cite{Marion1989}. In section \ref{sec:KSE} we study the best approach to organizing the eigenvectors. 

\ALrevise{With Eq.\ \ref{eq:resolved} and Eq.\ \ref{eq:unresolved}} ROM can be performed by either assuming $q=0$, which is the standard Galerkin approach, or by assuming $\mathrm{d}q/\mathrm{d}t=0$, which is the nonlinear Galerkin approach \cite{Marion1989,Jauberteau1990}. With the standard Galerkin approach Eq.\ \ref{eq:resolved} can be solved forward in time without any additional equations. With the nonlinear Galerkin approach $q$ is \ALrevise{approximated iteratively by} 
\begin{equation}\label{eq:q}
	q_{i+1}= \tilde{A}^{-1} Q \tilde{F}(p+q_i),
\end{equation}
\ALrevise{ with an initial guess of $q_0=0$. In our trials we only use one iteration to calculate $q_1$, because additional iterations had little affect on the results.}
Nonlinear Galerkin is more accurate than Galerkin, but is computationally more expensive due to solving for $q$ at every timestep. Another alternative is postprocessing Galerkin where $p$ is solved forward in time using standard Galerkin and $q$ is approximated for all $p$ at the end with Eq.\ \ref{eq:q} \cite{GarciaArchilla1998}. 

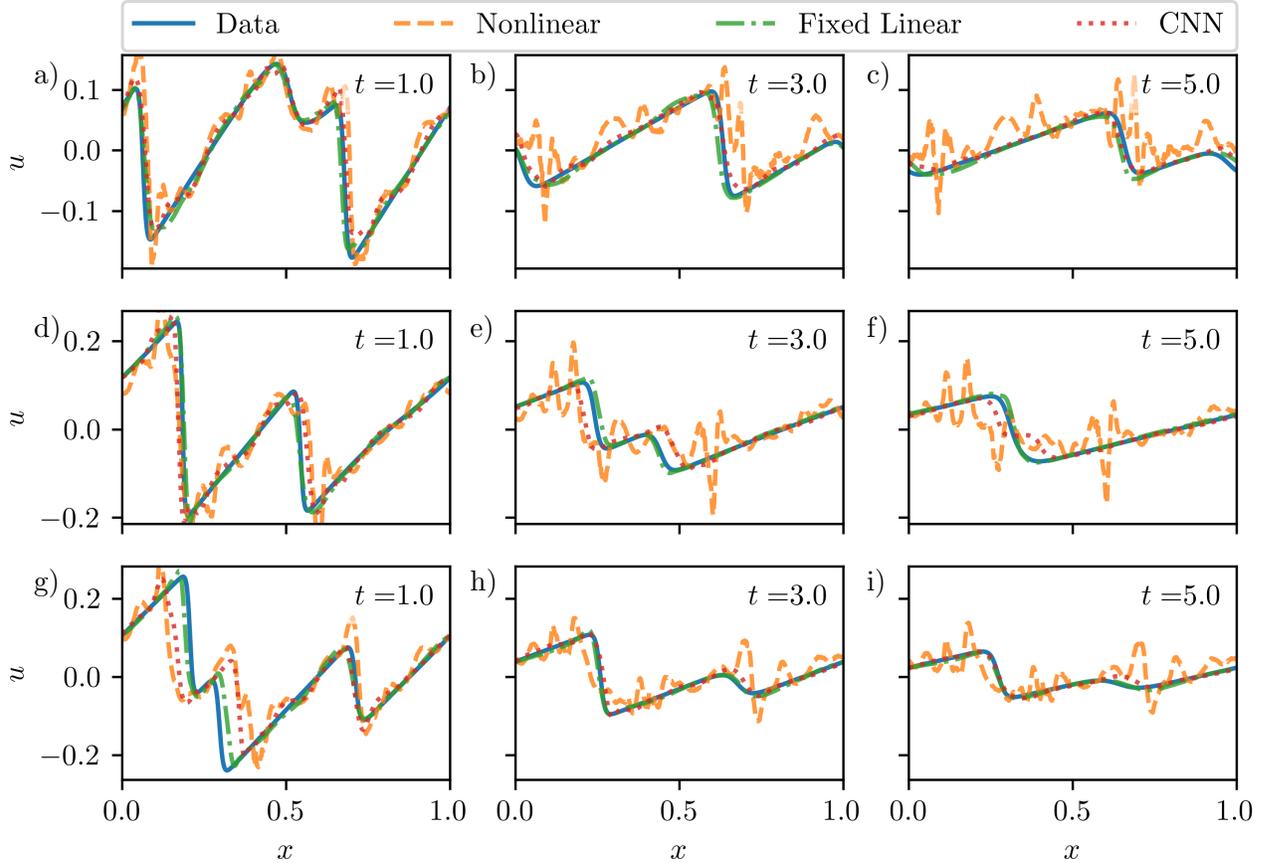
\begin{figure*}
	\centering
	\captionsetup[subfigure]{labelformat=empty}
	\begin{subfigure}[b]{17.2 cm}
		\includegraphics[trim=0 0 0 0,width=\textwidth,clip]{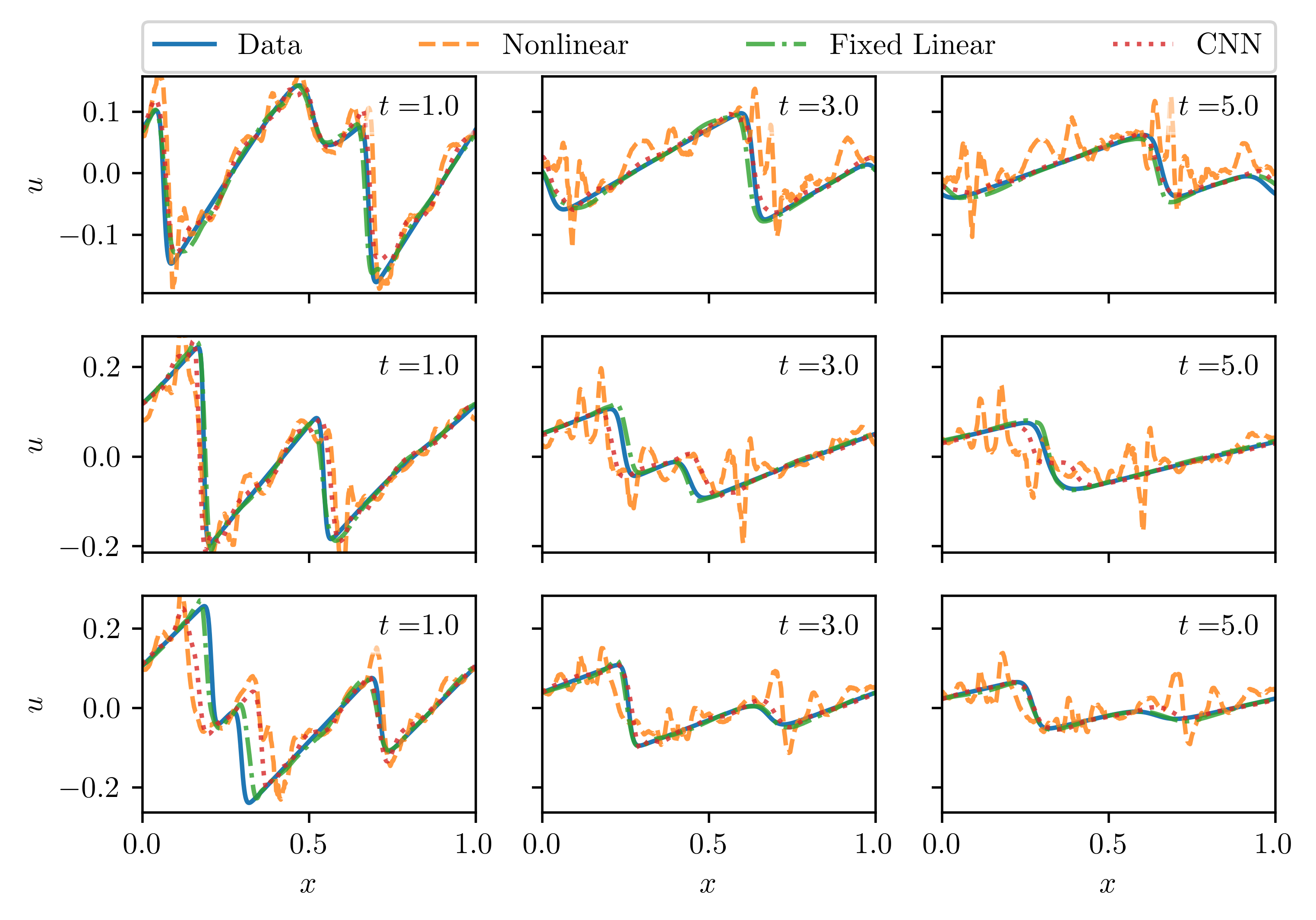}
		\begin{picture}(0,0)
		\put(-225,325){a)}
		\put(-60,325){b)}
		\put(90,325){c)}
		\put(-225,229){d)}
		\put(-60,229){e)}
		\put(90,229){f)}
		\put(-225,133){g)}
		\put(-60,133){h)}
		\put(90,133){i)}
		\end{picture}
		\caption{}
		\vspace{-10mm}
		\label{fig:Traj11}
	\end{subfigure}
	\begin{subfigure}[b]{0\textwidth}\caption{}\vspace{-10mm}\label{fig:Traj13}\end{subfigure}
	\begin{subfigure}[b]{0\textwidth}\caption{}\vspace{-10mm}\label{fig:Traj15}\end{subfigure}
	\begin{subfigure}[b]{0\textwidth}\caption{}\vspace{-10mm}\label{fig:Traj21}\end{subfigure}
	\begin{subfigure}[b]{0\textwidth}\caption{}\vspace{-10mm}\label{fig:Traj23}\end{subfigure}
	\begin{subfigure}[b]{0\textwidth}\caption{}\vspace{-10mm}\label{fig:Traj25}\end{subfigure}
	\begin{subfigure}[b]{0\textwidth}\caption{}\vspace{-10mm}\label{fig:Traj31}\end{subfigure}
	\begin{subfigure}[b]{0\textwidth}\caption{}\vspace{-10mm}\label{fig:Traj32}\end{subfigure}
	\begin{subfigure}[b]{0\textwidth}\caption{}\vspace{-10mm}\label{fig:Traj35}\end{subfigure}
	\captionsetup{justification=raggedright}
	\caption{Snapshots of \ALrevise{VBE} trajectories of true system and model predictions for different initial conditions. The rows indicate different initial conditions and the columns indicate solution fields at $t=1$, $3$, and $5$, respectively.}
	\label{fig:Traj}
	\vspace{-5mm}
\end{figure*}

\section{Results} \label{sec:Results}

In the following sections we demonstrate the importance of constructing neural ODEs with a linear and a nonlinear term on the VBE and the KSE. We examine the VBE to highlight the ability of this method to accurately capture the shock behavior seen at low viscosity and to accurately capture the energy spectrum. Then, we consider the KSE to show how our approach is needed to keep the long-time trajectories of a chaotic dynamical system on the attractor. This then leads to a demonstration of reduced order modeling via Galerkin and nonlinear Galerkin using the eigenvectors of the linear term we discover. In all cases results are shown for test data that was not used for training any of the models. Neural ODEs are trained using PyTorch \cite{Paszke2019} with code modified from Chen et al.\ \cite{Chen2019}.

\subsection{Viscous Burgers Equation} \label{sec:VBE}

In this section the data comes from numerical solutions of the 1D viscous Burgers equation
\begin{equation}\label{eq:Burgers}
	\dfrac{\partial v}{\partial t}=-v\dfrac{\partial v}{\partial x}+\nu\dfrac{\partial^2 v}{\partial x^2}, 
\end{equation}
in a domain of unit length ($L=1$) with periodic boundary conditions, with viscosity $\nu=8 \cdot 10^{-4}$. \ALrevise{This parameter set results in what is known as Burgers turbulence or \emph{Burgulence} \cite{MAULIK2020}.}
We solve this equation using a pseudospectral Runge-Kutta Crank-Nicolson scheme as described in \cite{san2013stationary}. The state $u$ is then represented in $\mathbb{R}^{512}$ by $v$ sampled at equidistant points in the domain. 
\ALrevise{We construct 1000 random initial conditions for training the models and 100 random initial conditions for testing model performance. We then evolve each initial condition over the interval $[0,5]$ sampling every $0.05$ time units. This results in $10^4$ total snapshots of data used for training.} \ALrevise{All figures compare test data to model reconstruction of the test data.} 

\ALrevise{When generating training and test datasets, the space of feasible initial conditions is infinite, so we must first limit this region. Here, we chose to limit the random initial conditions such that the ensemble-averaged energy spectrum}
\begin{equation}\label{eq:KE}
	E(k,t)=\left<\dfrac{1}{2} |\hat{u}(k,t)|^2\right>
\end{equation}
\ALrevise{ at $t=0$ satisfies $\int E(k,0) \, dk=0.5L/(2\pi)$. In Eq.\ \ref{eq:KE} $k$ is the wavenumber and the Fourier transform is $\hat{u}=\mathcal{F}(u)$. This energy condition is satisfied by selecting the initial conditions according to} 
\begin{equation}
    \hat{u}(k,0)=\sqrt{2 E_0(k)} \left(\cos( 2 \pi \Psi(k))-\sin(2 \pi \Psi(k)\right),
\end{equation}
\ALrevise{with $E_0(k)=A k^{4} \exp \left(-\left(k / k_{0}\right)^{2}\right)$, and $\Psi(k)$ coming from a uniform distribution between 0 and 1 for each wavenumber ($\Psi(k)=\mathcal{U}(0,1)$). The constants $A$ and $k_0$ are the same as in \cite{MAULIK2020}. Selecting the initial conditions in this manner results in training and testing datasets with a variety of different initial energies.}



\begin{table}
	\captionsetup{justification=raggedright}
	\caption{Architectures of NNs and matrices used in Sections \ref{sec:VBE}. ``Shape" indicates the dimension of each layer, ``Activation" the corresponding activation functions, ``Learning Rate" is the learning rate the NN switches to evenly over the training period, and ``Weight Init" is the distribution used to initialize the weights of the NNs. \ALrevise{``relu" is the rectified linear unit \cite{IanGoodfellowYoshuaBengio2017}.} $\tilde{F}_1$ is the NN trained with the true $A$ and $\tilde{F}_2$ is the NN trained with $\tilde{A}$.}
	\resizebox{.81\textwidth}{!}{%
		\begin{tabular}{l*{6}{c}r}
			Function & Shape & Activation & Learning Rate & Weight Init \\
			\hline
			$\tilde{f}$		& 512/200/200/200/512 \quad & relu/relu/relu/linear & $[10^{-3},10^{-4},10^{-5}]$ 	& $\mathcal{N}(0,10^{-2})$\\
			$\tilde{F}_1$	& 512/200/200/200/512 \quad & relu/relu/relu/linear & $[10^{-3},10^{-4},10^{-5}]$ 	& $\mathcal{N}(0,10^{-2})$\\
			$\tilde{F}_2$	& 512/200/200/200/512 \quad & relu/relu/relu/linear & $[10^{-3},10^{-4}]$			& $\mathcal{N}(0,10^{-2})$\\
			$\tilde{A}$		& 512/512 				& linear 				& $[10^{0},10^{-1},10^{-2}]$  	& $\mathcal{N}(0,10^{4})$\\
		\label{Table}
		\end{tabular}}
\end{table}

With this data, we train neural ODEs for \ALrevise{the three cases listed above: 1) $\tilde{h}(u)=\tilde{f}(u;\theta_1)$, 2) $\tilde{h}(u)= Au+\tilde{F}(u;\theta_2)$, and 3) $\tilde{h}(u)=\tilde{A}(\theta_3)u+\tilde{F}(u;\theta_2)$, which we will denote as nonlinear, fixed linear, and CNN. We train these models} 
for $10^4$ epochs at which point the error \ALrevise{in the loss (Eq. \ref{eq:ODENet}) stops decreasing.}
The architectures and training parameters for each of these NNs are presented in Table \ref{Table}. \ALrevise{Hyperparameter tuning was performed manually by training many NNs and selecting the NNs with the best short-time tracking performance. The linear term in particular showed sensitivity to hyperparemeter selection, which should be investigated further in future work.}

We construct $\tilde{f}$ and $\tilde{F}$ from fully connected NNs, and $\tilde{A}$ from a linear CNN with stride one, one filter, and a filter width of 3. This is equivalent to learning a sparse tridiagonal matrix. \ALrevise{In the case of the VBE,} we enforce symmetry \ALrevise{in $\tilde{A}$ by training a CNN $B(\theta_3)$ and letting $\tilde{A}=B(\theta_3)+B(\theta_3)^T$.} 
With this tridiagonal structure, the optimal filter for approximating the derivative is $[210,-420,210]$ \ALrevise{(this is the approximation of the second derivative with central differencing)}, and the filter learned after training is $[45,-90,45]$\ALrevise{. Although the scale of the filter is off, we will see the results with this filter and the true filter match well.}

In Fig. \ref{fig:Traj} we show the performance of \ALrevise{the three neural ODE} approaches on different initial conditions of the VBE. For all three initial conditions we see that \ALrevise{stabilized neural ODE approaches (fixed linear and CNN)}
result in good agreement between the true solution and the model prediction. When, instead \ALrevise{we take the nonlinear approach}
high-wavenumber behavior appears in the solution and persists at long-times resulting in a poor prediction.

\begin{figure} 
    \centering
	\includegraphics[trim=0 0 0 0,width=8.6 cm,clip]{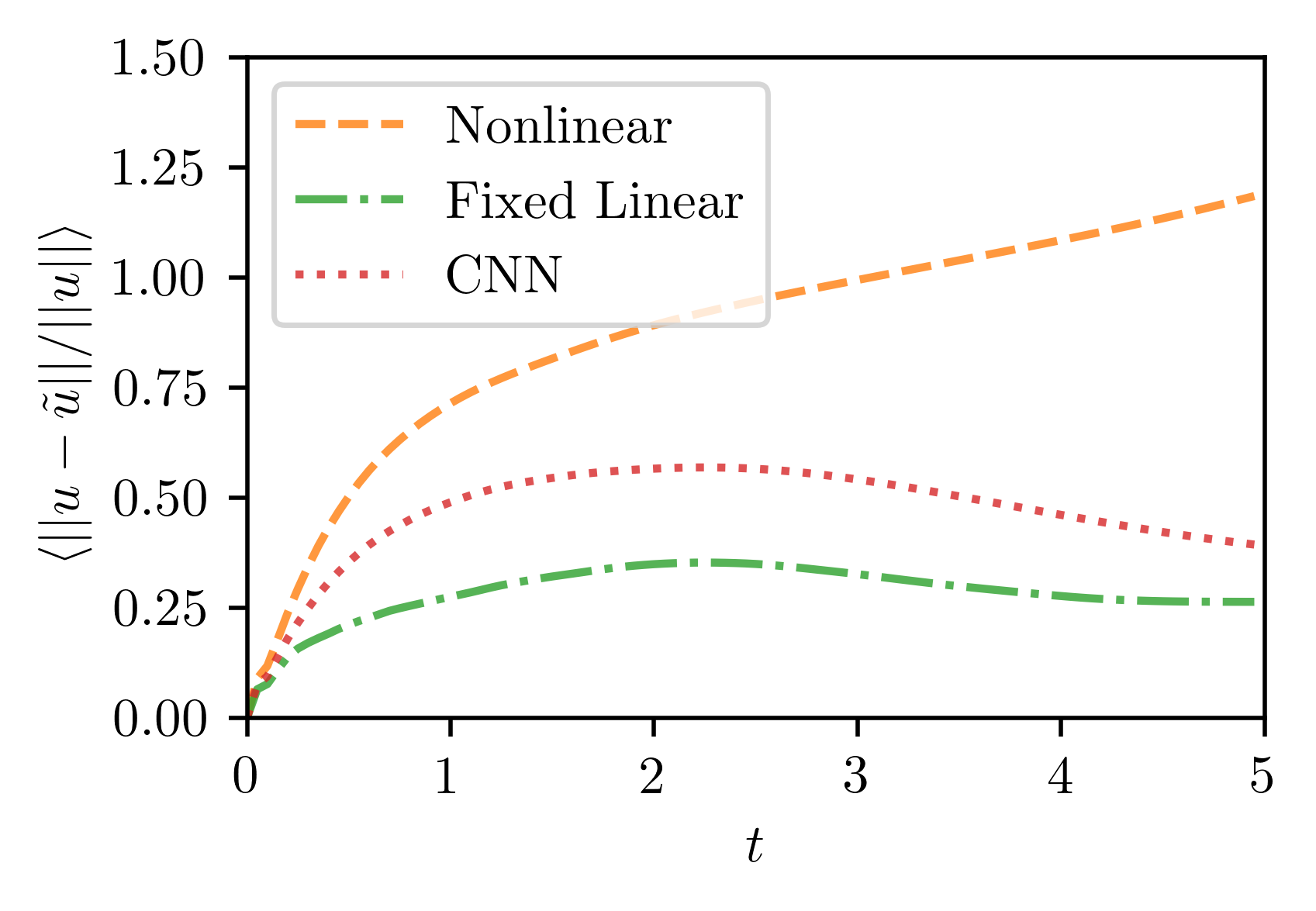}
	\caption{Ensemble averaged error of model predictions VBE.}
	\label{fig:Burgers_Dif}
\end{figure} 

The behavior just described becomes even more pronounced when considering the relative error averaged over initial conditions, shown in Fig.\ \ref{fig:Burgers_Dif}. Here, all methods match at very short times. The fixed linear approach performs the best followed by the CNN approach, and then by the nonlinear approach. For the two \ALrevise{stabilized neural ODE} methods the errors level off, or slightly go down, at long-times indicating that trajectories tend to separate somewhat and then decay in a similar fashion causing them to stay the same relative distance apart. However, this is not the case for the nonlinear method where the high-wavenumber behavior at long-times continues to push the trajectory further away increasing the relative error.

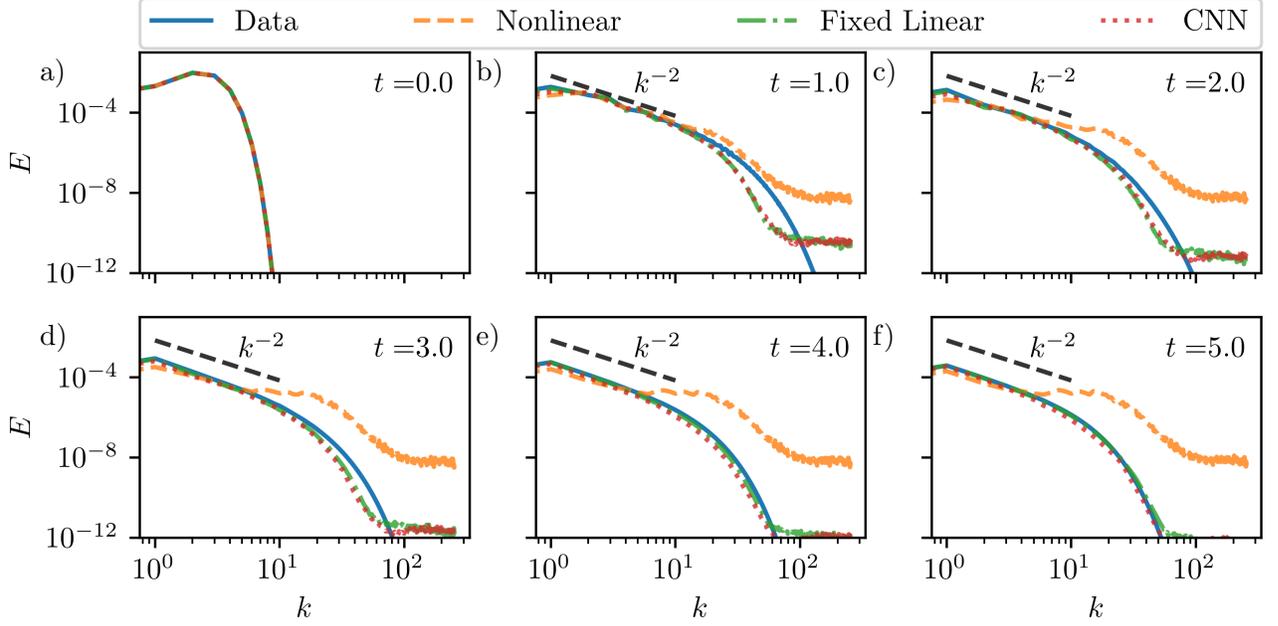
\begin{figure*}
	\centering
	\captionsetup[subfigure]{labelformat=empty}
	\begin{subfigure}[b]{17.2 cm}
		\includegraphics[trim=0 0 0 0,width=\textwidth,clip]{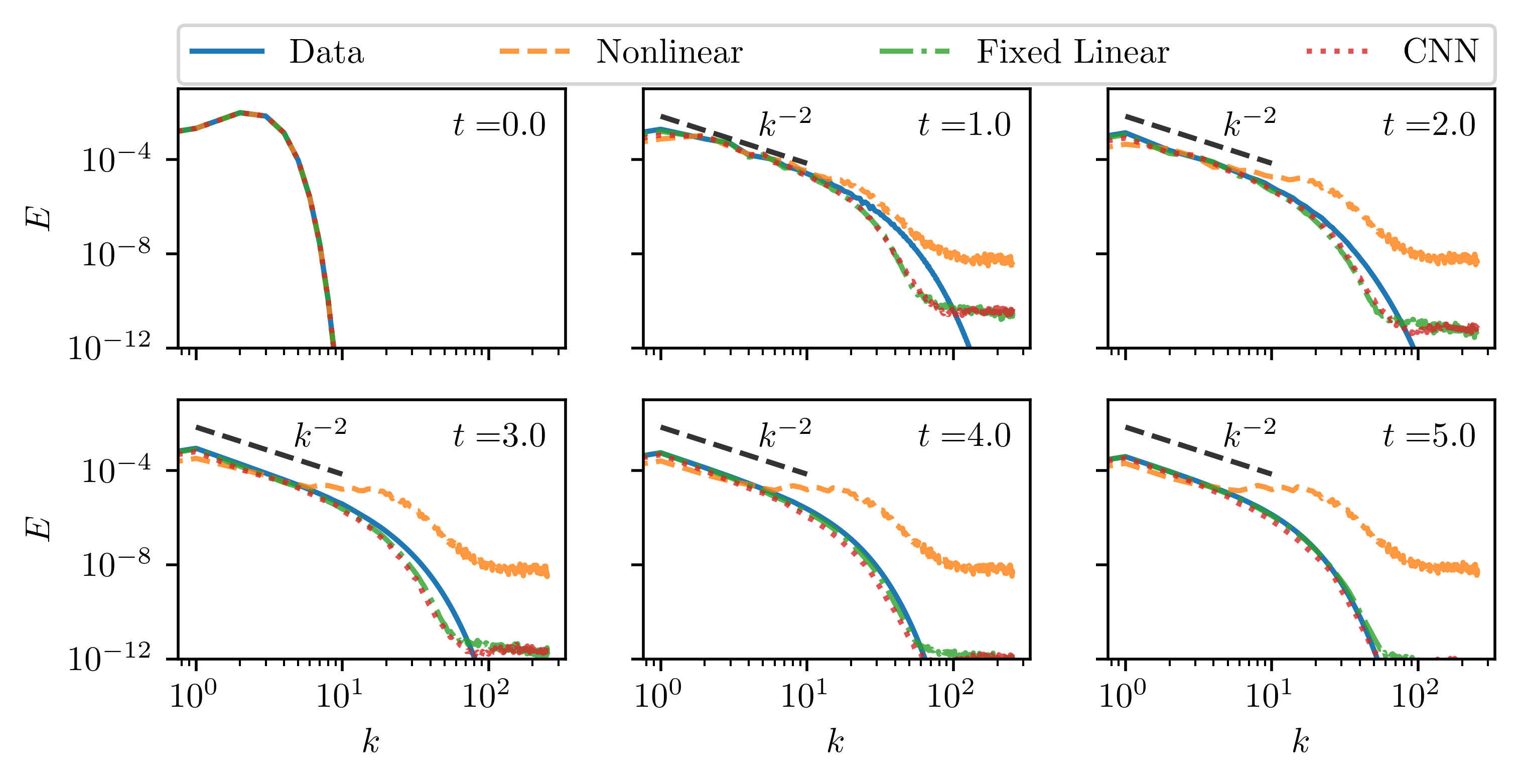}
		\begin{picture}(0,0)
		\put(-225,235){a)}
		\put(-60,235){b)}
		\put(90,235){c)}
		\put(-225,135){d)}
		\put(-60,135){e)}
		\put(90,135){f)}
		\end{picture}
		\caption{}
		\vspace{-10mm}
		\label{fig:PSD0}
	\end{subfigure}
	\begin{subfigure}[b]{0\textwidth}\caption{}\vspace{-10mm}\label{fig:PSD1}\end{subfigure}
	\begin{subfigure}[b]{0\textwidth}\caption{}\vspace{-10mm}\label{fig:PSD2}\end{subfigure}
	\begin{subfigure}[b]{0\textwidth}\caption{}\vspace{-10mm}\label{fig:PSD3}\end{subfigure}
	\begin{subfigure}[b]{0\textwidth}\caption{}\vspace{-10mm}\label{fig:PSD4}\end{subfigure}
	\begin{subfigure}[b]{0\textwidth}\caption{}\vspace{-10mm}\label{fig:PSD5}\end{subfigure}
	\captionsetup{justification=raggedright}
	\caption{Ensemble-averaged energy spectrum of the true system and the model predictions at different times \ALrevise{, with $k^{-2}$ scaling for reference}. (a)-(f) are between $t=0-5$.}
	\label{fig:PSD}
	\vspace{-5mm}
\end{figure*}   

\ALrevise{We also consider the ensemble-averaged energy spectrum defined in Eq.\ \ref{eq:KE}.}
Fig.\ \ref{fig:PSD} shows the energy spectrum at various times. At early times, when the shock develops, there is an increase in the tail of the energy spectrum, and we see the development of a $k^{-2}$ scaling. At $t=1$ \ALrevise{all} methods capture most of the spectrum with errors appearing at high-wavenumbers. Then, at $t\gtrsim2$, both \ALrevise{the stabilized methods} match the true energy spectrum over \ALrevise{full range of} wavenumbers, while the high-wavenumber behavior of the nonlinear model deviates drastically \ALrevise{as time increases}. Capturing the dynamics properly at all scales requires a linear term to damp out high wavenumbers. 

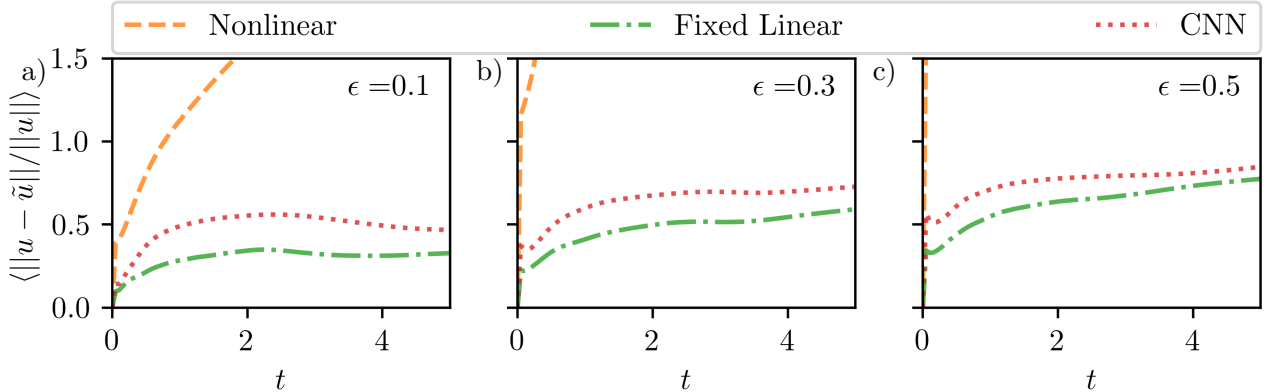
\begin{figure*}
	\centering
	\captionsetup[subfigure]{labelformat=empty}
	\begin{subfigure}[b]{17.2 cm}
		\includegraphics[trim=0 0 0 0,width=\textwidth,clip]{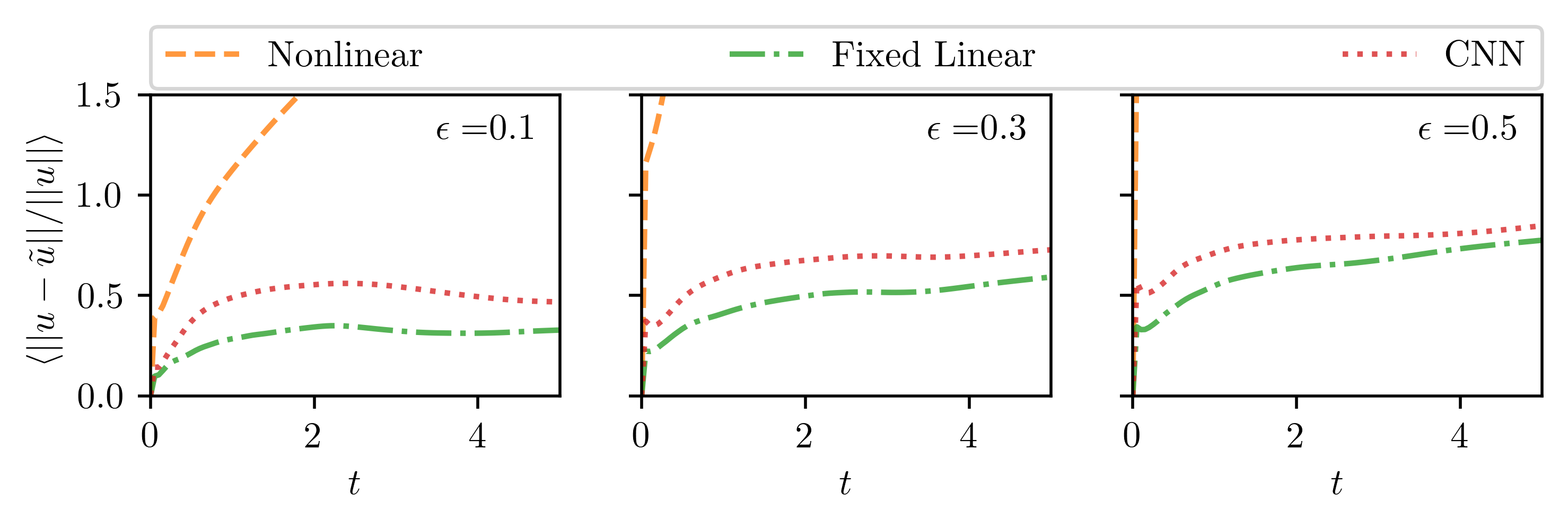}
		\begin{picture}(0,0)
		\put(-232,149){a)}
		\put(-60,149){b)}
		\put(90,149){c)}
		\end{picture}
		\caption{}
		\vspace{-10mm}
		\label{fig:Burgers_Dif_Noise1}
	\end{subfigure}
	\begin{subfigure}[b]{0\textwidth}\caption{}\vspace{-10mm}\label{fig:Burgers_Dif_Noise3}\end{subfigure}
	\begin{subfigure}[b]{0\textwidth}\caption{}\vspace{-10mm}\label{fig:Burgers_Dif_Noise5}\end{subfigure}
	\captionsetup{justification=raggedright}
	\caption{Ensemble averaged error of predictions with noisy initial conditions of $\epsilon=0.1$, $0.3$, and $0.5$ for (a)-(c).}
	\label{fig:Burgers_Dif_Noise}
	\vspace{-5mm}
\end{figure*}   

The \ALrevise{explicit} linear term \ALrevise{in our approach} also contributes to the robustness and generalizability of the neural ODE setup. Figure \ref{fig:Burgers_Dif_Noise} shows the relative error in predictions for various levels of noise added to the initial condition. To create this figure, we add random Gaussian noise of the form $\mathcal{N}(0,\epsilon^2)$ to each grid location for 100 different initial conditions. Then those initial conditions are evolved forward with the true equations and the learned models for comparison. In all these cases the error in the nonlinear cases rapidly increases, while the \ALrevise{stabilized neural ODE} cases exhibit similar behavior as with no noise. This shows that these models generalize well to noisy initial conditions that are much different from the training data. This can also be seen in Fig.\ \ref{fig:Burgers_Dif_Noise1}, where we plot the ensemble averaged energy spectrum at different times for the different levels of noise. The \ALrevise{strictly} nonlinear model does a poor job of capturing the spectrum at any wavenumber, while the \ALrevise{stabilized neural ODEs} accurately capture the spectrum like they did for the initial conditions with no noise.

\begin{figure*}
	\centering
	\captionsetup[subfigure]{labelformat=empty}
	\begin{subfigure}[b]{17.2 cm}
		\includegraphics[trim=0 0 0 0,width=\textwidth,clip]{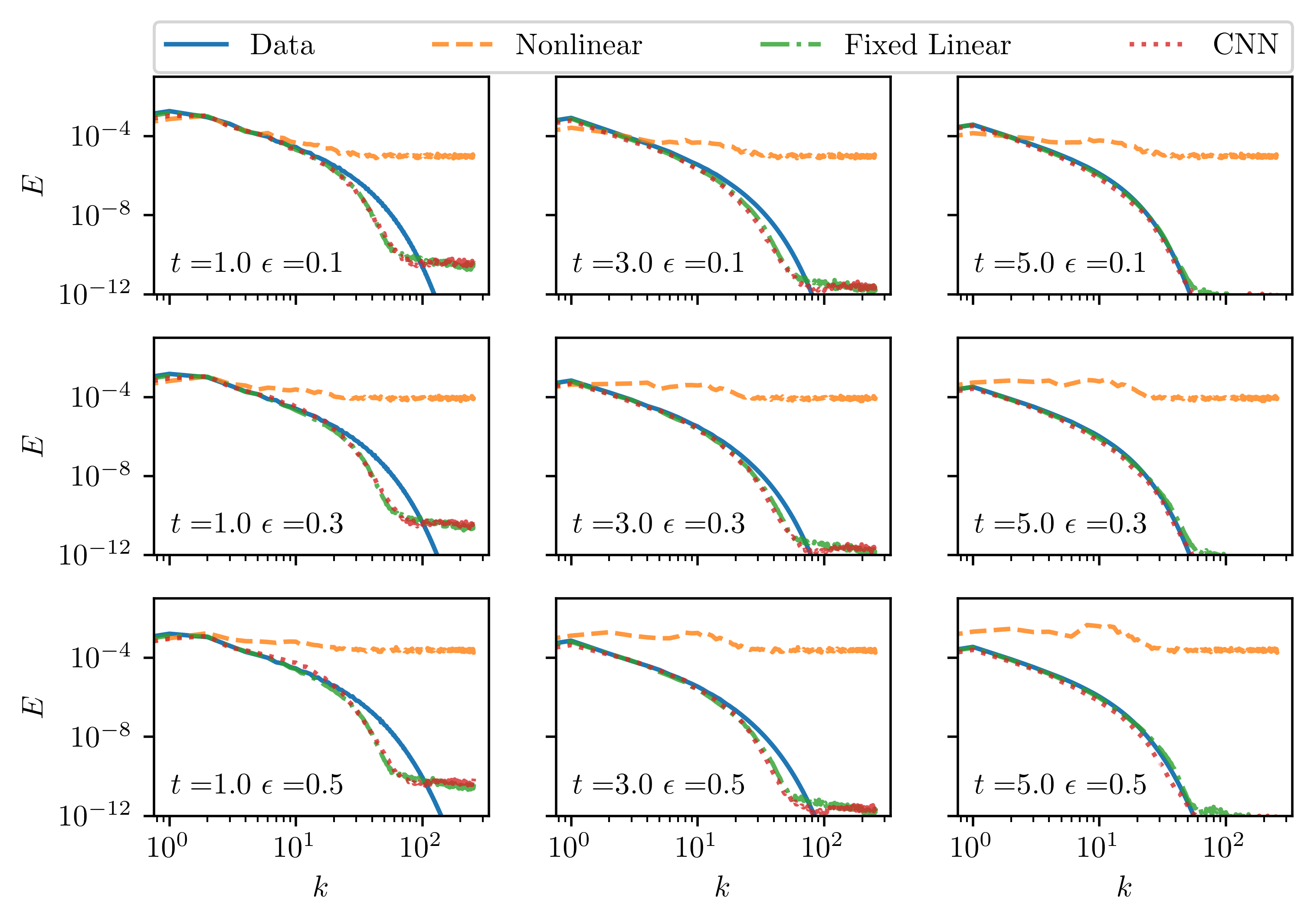}
		\begin{picture}(0,0)
		\put(-200,325){a)}
		\put(-50,325){b)}
		\put(100,325){c)}
		\put(-200,229){d)}
		\put(-50,229){e)}
		\put(100,229){f)}
		\put(-200,133){g)}
		\put(-50,133){h)}
		\put(100,133){i)}
		\end{picture}
		\caption{}
		\vspace{-10mm}
		\label{fig:PSD_Noise11}
	\end{subfigure}
	\begin{subfigure}[b]{0\textwidth}\caption{}\vspace{-10mm}\label{fig:PSD_Noise13}\end{subfigure}
	\begin{subfigure}[b]{0\textwidth}\caption{}\vspace{-10mm}\label{fig:PSD_Noisej15}\end{subfigure}
	\begin{subfigure}[b]{0\textwidth}\caption{}\vspace{-10mm}\label{fig:PSD_Noise21}\end{subfigure}
	\begin{subfigure}[b]{0\textwidth}\caption{}\vspace{-10mm}\label{fig:PSD_Noise23}\end{subfigure}
	\begin{subfigure}[b]{0\textwidth}\caption{}\vspace{-10mm}\label{fig:PSD_Noise25}\end{subfigure}
	\begin{subfigure}[b]{0\textwidth}\caption{}\vspace{-10mm}\label{fig:PSD_Noise31}\end{subfigure}
	\begin{subfigure}[b]{0\textwidth}\caption{}\vspace{-10mm}\label{fig:PSD_Noise32}\end{subfigure}
	\begin{subfigure}[b]{0\textwidth}\caption{}\vspace{-10mm}\label{fig:PSD_Noise35}\end{subfigure}
	\captionsetup{justification=raggedright}
	\caption{Ensemble-averaged energy spectrum of the true system and the model predictions at different times and noise on the initial conditions. (a)-(c),(d)-(f), and (g)-(i) are with noise $\epsilon=0.1$, $0.3$, and $0.5$, at times $t=1$, $3$, and $5$, respectively.}
	\label{fig:PSD_Noise}
	\vspace{-5mm}
\end{figure*}
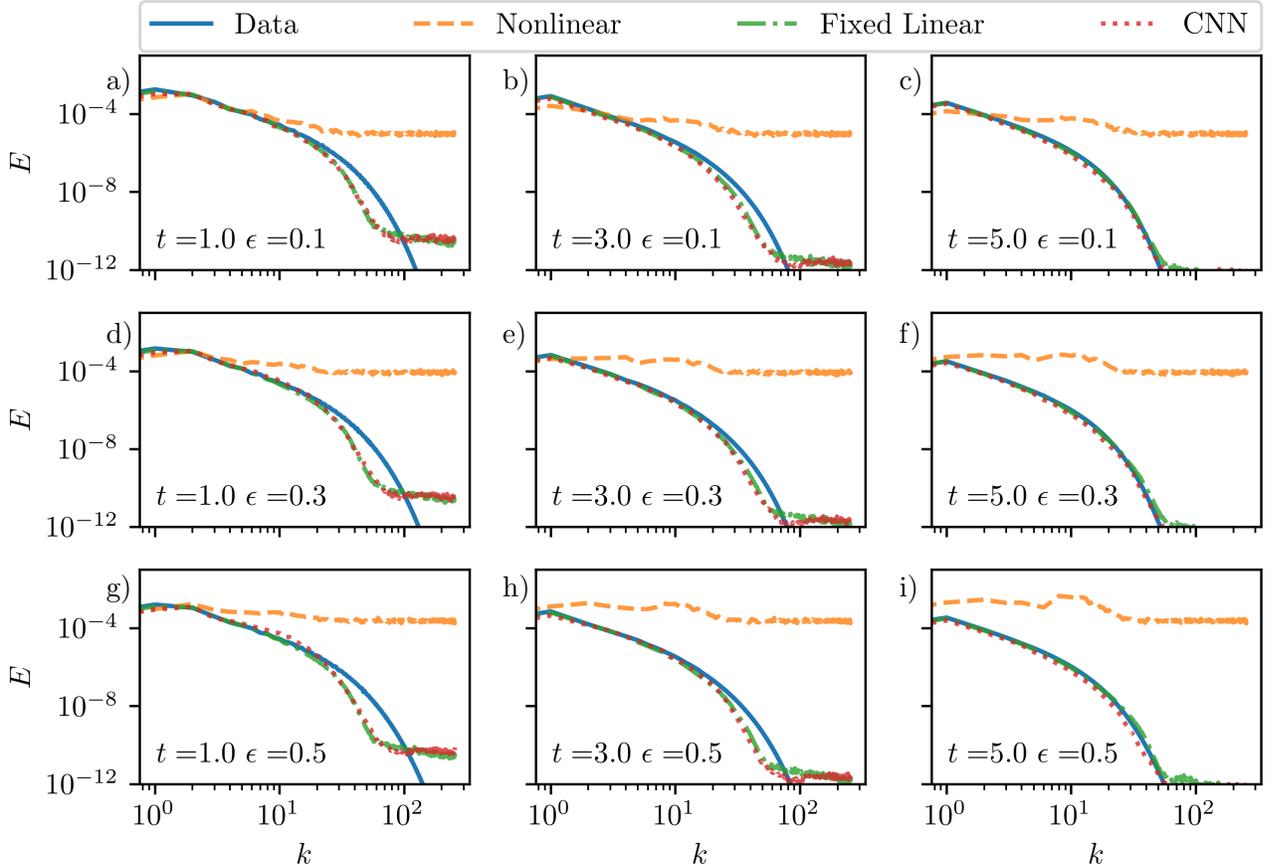   


\subsection{Kuramoto-Sivashinsky Equation} \label{sec:KSE}

In the previous section we showed that our setup captures the shock dynamics and transient decay exhibited by the VBE. In that case, all solutions decay to zero at long times.
Now we examine the KSE, where the dynamics are chaotic and collapse onto a low-dimensional inertial manifold at long-times \cite{Foias1988a,Temam1994,Jolly2000,Zelik2014}. A useful model for the KSE should keep long-time trajectories on the attractor and provide a means for reduced-order modeling to capitalize on the low-dimensionality of the manifold on which the data lies.

We numerically solve the 1D KSE,
\begin{equation}\label{eq:KSE}
	\dfrac{\partial v}{\partial t}=-v\dfrac{\partial v}{\partial x}-\dfrac{\partial^2 v}{\partial x^2}-\dfrac{\partial^4 v}{\partial x^4}, 
\end{equation}
with periodic boundary conditions on a domain of size $L=22$. Solutions are found by performing a Galerkin projection onto Fourier modes and using exponential time differencing to evolve the ODE forward in time \cite{Kassam2005}. After solving in Fourier space, we then transform the data back to physical space for training the neural ODEs. Code for solving the KSE is available in Cvitanovi\'c et al.\ \cite{ChaosBook}. The data used is a single trajectory that has collapsed onto the inertial manifold evolved over $t=[0,10^5]$ sampled every $0.25$ \ALrevise{time units}. \ALrevise{The first 80\% of this data is used for training and the remaining 20\% is used as test data.}

\begin{table}
	\captionsetup{justification=raggedright}
	\caption{Neural ODE details for Section \ref{sec:KSE}. Labels are the same as in Table \ref{Table}. ``Sig" is the sigmoid activation \cite{IanGoodfellowYoshuaBengio2017}. 
	}
	\resizebox{.81\textwidth}{!}{%
		\begin{tabular}{l*{6}{c}r}
			Function & Shape & Activation & Learning Rate & Weight Init \\
			\hline
			$\tilde{f}$		& 64/200/200/200/64 	\quad & sig/sig/sig/linear 	& $[10^{-3},10^{-4}]$ 			& $\mathcal{N}(0,10^{-2})$\\
			$\tilde{F}_1$	& 64/200/200/200/64  & sig/sig/sig/linear 	& $[10^{-3},10^{-4}]$ 			& $\mathcal{N}(0,10^{-2})$\\
			$\tilde{F}_2$	& 64/200/200/200/64  & sig/sig/sig/linear 	& $[10^{-3},10^{-4}]$			& $\mathcal{N}(0,10^{-2})$\\
			$\tilde{A}$		& 64/64 					& linear 				& $[10^{0},10^{-1},10^{-2}]$  	& $\mathcal{U}(-\sqrt{1/3},\sqrt{1/3})$\\
		\label{Table2}
		\end{tabular}}
\end{table}

We train neural ODEs for $4\cdot 10^4$ epochs, \ALrevise{and consider the same three cases as before.} Table \ref{Table2} contains details on the architectures and training parameters used for each of these NNs. \ALrevise{Small modifications were made in the architectures due to the different dynamics of the KSE.} As before, $\tilde{f}$ and $\tilde{F}$ are fully connected NN and $\tilde{A}$ is a linear CNN. For $\tilde{A}$ we selected a wider filter of width 5 because higher-order gradients need to be approximated. In general the linear term is unknown, so the filter width can be viewed as an additional tuning parameter that is similar to selecting the stencil size for a finite difference approximation. For our system, and filter width, the optimal filter parameters are $[-72,278,-413,278,-72]$ \ALrevise{(this is central differencing of the diffusive and hyperdiffusive terms)}, while the parameters learned for the CNN were $[-0.7,1.8,-2.7,1.8,-0.7]$. \ALrevise{Here we also ran trials where we forced symmetry in the filter, but we found the training procedure resulted in the symmetry without the need to enforce it.}
\ALrevise{Despite} the magnitude of the parameters being much smaller, we will show the CNN successfully damps out high wavenumbers\ALrevise{, and gives the same eigenvectors}.


 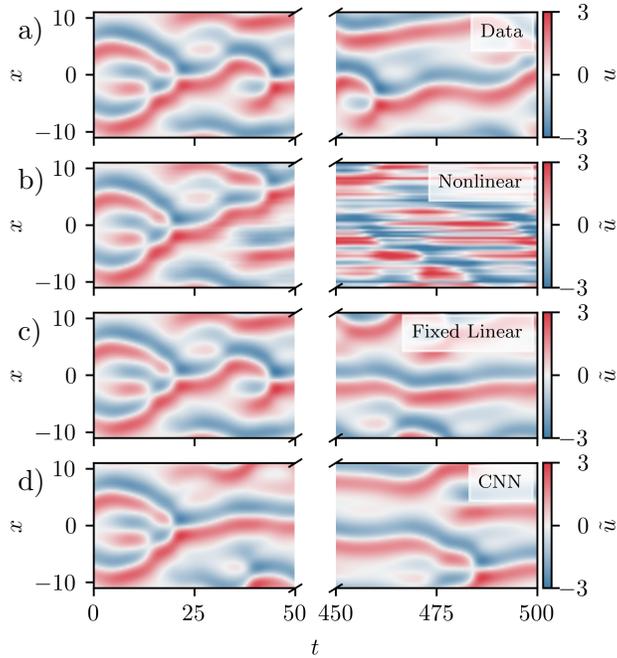
\begin{figure}
	\centering
	\captionsetup[subfigure]{labelformat=empty}
	\begin{subfigure}[b]{8.6 cm}
		\includegraphics[trim=0 0 0 0,width=\textwidth,clip]{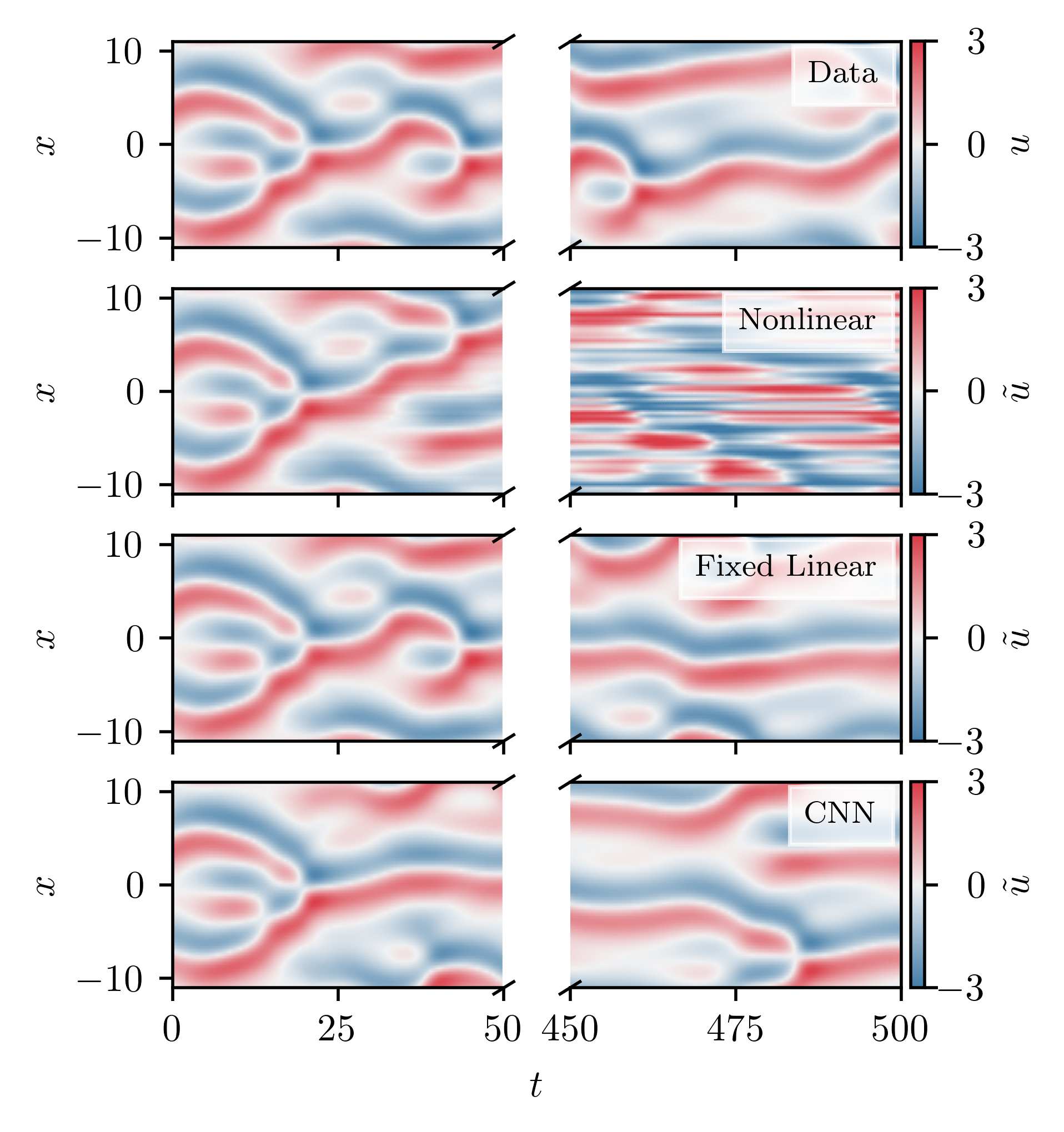}
		\begin{picture}(0,0)
		\put(-115,263){\contour{white}{ \textcolor{black}{a)}}}
		\put(-115,206){\contour{white}{ \textcolor{black}{b)}}}
		\put(-115,149){\contour{white}{ \textcolor{black}{c)}}}
		\put(-115,92){\contour{white}{ \textcolor{black}{d)}}}
		\end{picture}
		\caption{}
		\vspace{-10mm}
		\label{fig:Traja}
	\end{subfigure}
	\begin{subfigure}[b]{0\textwidth}\caption{}\vspace{-10mm}\label{fig:Trajb}\end{subfigure}\begin{subfigure}[b]{0\textwidth}\caption{}\vspace{-10mm}\label{fig:Trajc}\end{subfigure}\begin{subfigure}[b]{0\textwidth}\caption{}\vspace{-10mm}\label{fig:Trajd}\end{subfigure}    
	\vspace{-1.5\baselineskip}
	\captionsetup{justification=raggedright}
	\caption{Predictions of trajectories using different models. (a) is the true trajectory, (b) is the predicted trajectory with the nonlinear model. (c) is the predicted trajectory with the fixed linear model. (d) is the predicted trajectory with the CNN model.}
	\label{fig:KSE_Traj} 
	\vspace{-5mm}
\end{figure}

Figure \ref{fig:KSE_Traj} compares a true trajectory to predictions from the three methods. The left side shows short-times, where model predictions should \ALrevise{match the true trajectory.} The right side shows long-times, where the model predictions should have statistics that agree with the true trajectory. In Fig. \ref{fig:Trajb} the \ALrevise{nonlinear method} prediction is shown. At early times the predicted trajectory matches the true trajectory, but at long-times \ALrevise{erroneous} high-wavenumber behavior appears and the trajectory has moved far away from the attractor. This result agrees with what was found in \cite{Linot2021}. Predictions from the \ALrevise{stabilized neural ODEs} appear in Figs. \ref{fig:Trajc} and \ref{fig:Trajd} for the fixed linear and CNN methods, respectively. In the linear cases, there is good agreement at short times, with the fixed linear method tracking slightly better, and at long-times the trajectories stay on the attractor. \ALrevise{Below we show statistics to further validate this conclusion.}

\begin{figure} 
    \centering
	\includegraphics[trim=0 0 0 0,width=8.6 cm,clip]{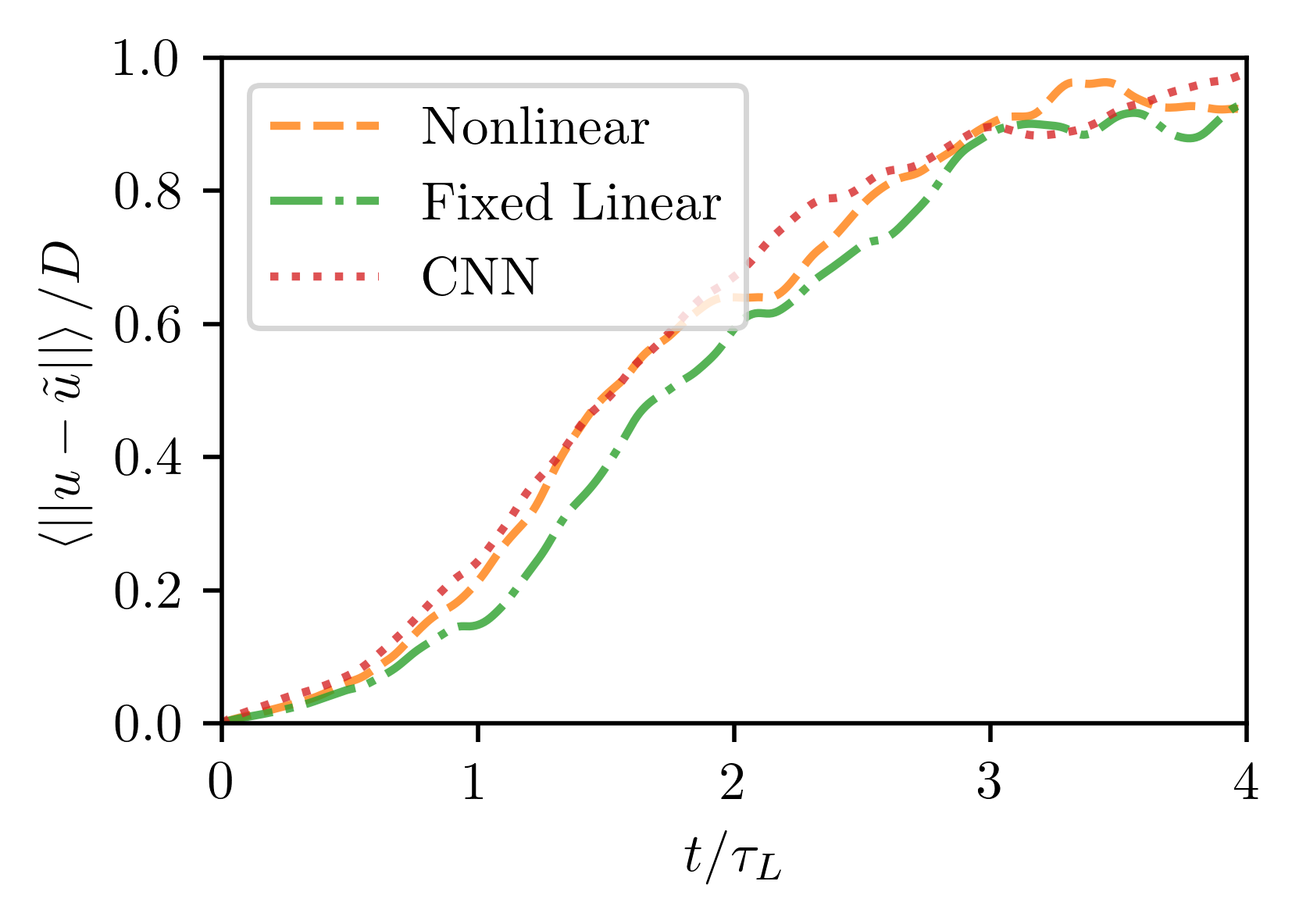}
	\caption{Ensemble averaged error of model predictions for the KSE.}
	\label{fig:KSE_Dif}
\end{figure} 


To better quantify the apparent results in Fig. \ref{fig:KSE_Traj} we consider some ensemble averaged quantities. In Fig.\ \ref{fig:KSE_Dif} we show the ensemble averaged error for 100 different initial conditions as a function of \ALrevise{Lyapunov} time \ALrevise{normalized by the difference between states on the attractor sampled at random times $t_i$ and $t_j$ ($D=\left< ||u(t_i)-u(t_j)||^2 \right>$)}. The Lyapunov time \ALrevise{is the inverse of the} leading Lyapunov exponent and is the relevant timescale for chaotic systems. For the KSE, with $L=22$, the Lyapunov time is $\tau_L\approx 22$. The error in all three models nearly match over this time range, with the fixed linear model performing slightly better. All three models are in agreement because the high-wavenumber behavior of the \ALrevise{strictly} nonlinear model does not dominate over these short time ranges. In all three cases there is good tracking, with trajectories completely diverging at around $t\sim 4 \tau_L$.  

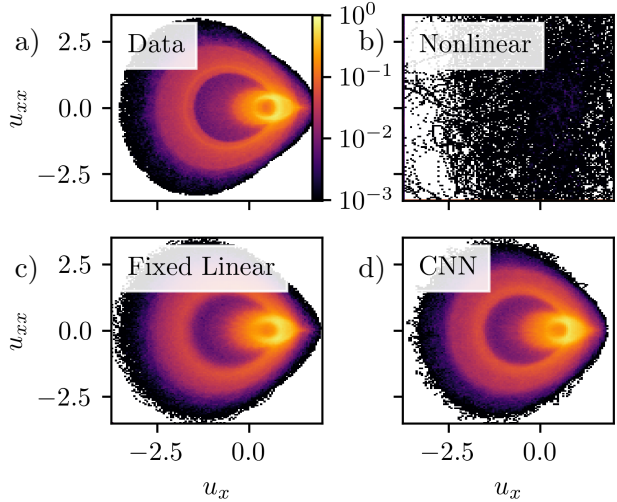
\begin{figure}
	\centering
	\captionsetup[subfigure]{labelformat=empty}
	\begin{subfigure}[b]{8.6 cm}
		\includegraphics[trim=0 0 0 0,width=8.6 cm,clip]{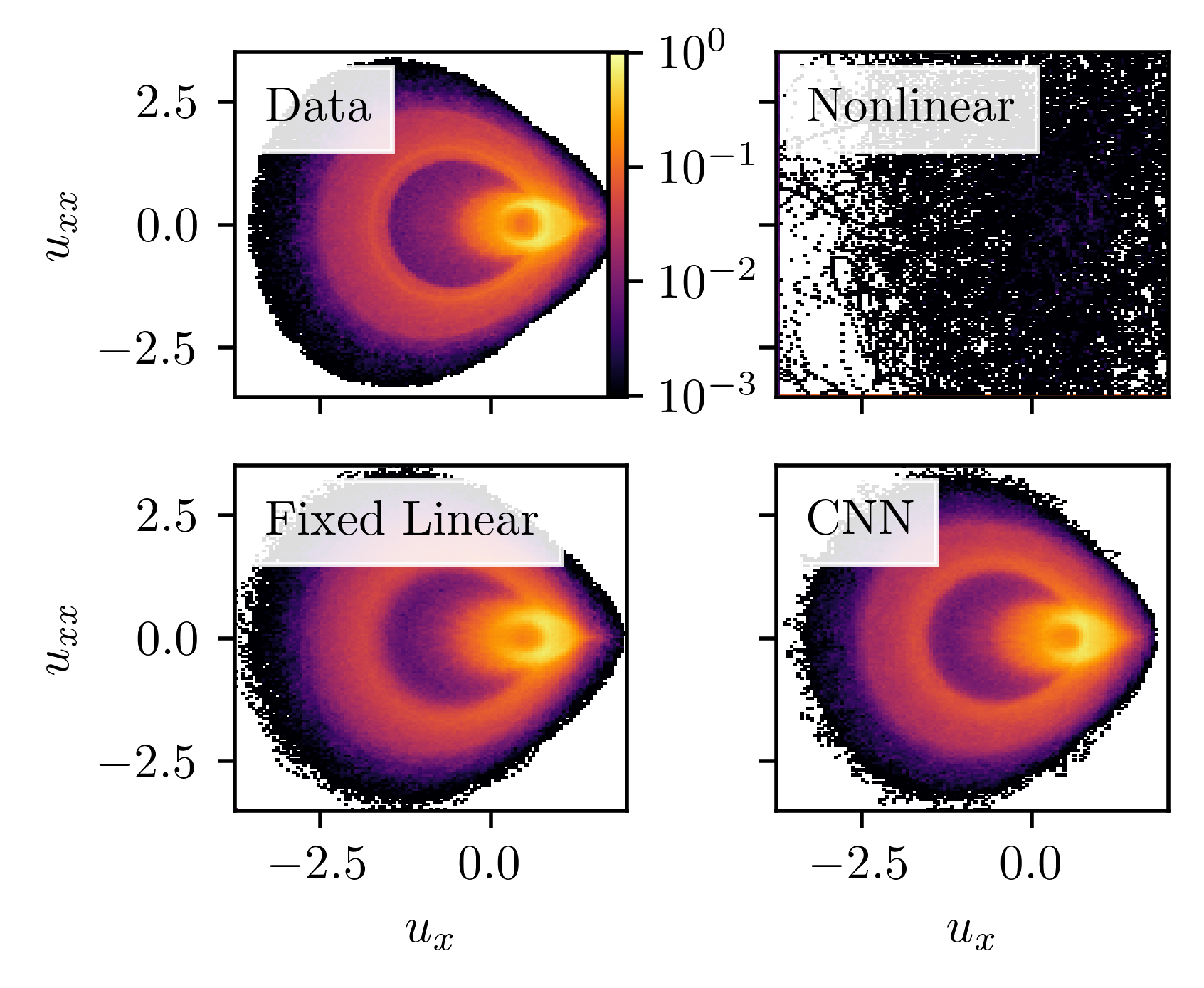}
		\begin{picture}(0,0)
		\put(-115,199){\contour{white}{ \textcolor{black}{a)}}}
		\put(15,199){\contour{white}{ \textcolor{black}{b)}}}
		\put(-115,113){\contour{white}{ \textcolor{black}{c)}}}
		\put(15,113){\contour{white}{ \textcolor{black}{d)}}}
		\end{picture}
		\caption{}
		\vspace{-10mm}
		\label{fig:PDFa}
	\end{subfigure}
	\begin{subfigure}[b]{0\textwidth}\caption{}\vspace{-10mm}\label{fig:PDFb}\end{subfigure}\begin{subfigure}[b]{0\textwidth}\caption{}\vspace{-10mm}\label{fig:PDFc}\end{subfigure}\begin{subfigure}[b]{0\textwidth}\caption{}\vspace{-10mm}\label{fig:PDFd}\end{subfigure}    
	\vspace{-1.5\baselineskip}
	\captionsetup{justification=raggedright}
    \caption{(a) is the true joint PDF at $L=22$. (b)-(d) are the joint PDFs for the nonlinear, fixed linear, and CNN models, respectively. }
	\label{fig:KSE_PDF} 
	\vspace{-5mm}
\end{figure}

The advantage \ALrevise{stabilized} neural ODEs have over the standard neural ODE\ALrevise{, for the KSE,} is the prediction at long times. In Fig.\ \ref{fig:KSE_PDF} we show the true joint PDF of the first ($u_x$) and second ($u_{xx}$) derivatives of the state plotted for every spatiotemporal point of a trajectory evolved forward for $10^4$ time units. Figure \ref{fig:PDFb} therein, which corresponds to the nonlinear ODE prediction, shows the probability of finding a point in the nonlinear simulations prediction with the correct first and second derivatives is nearly zero. Both \ALrevise{stabilized} neural ODEs, shown in Figs. \ref{fig:PDFc} and \ref{fig:PDFd}, accurately match the true joint PDF.

\begin{figure}
	\centering
	\captionsetup[subfigure]{labelformat=empty}
	\begin{subfigure}[b]{8.6 cm}
		\includegraphics[trim=0 0 0 0,width=\textwidth,clip]{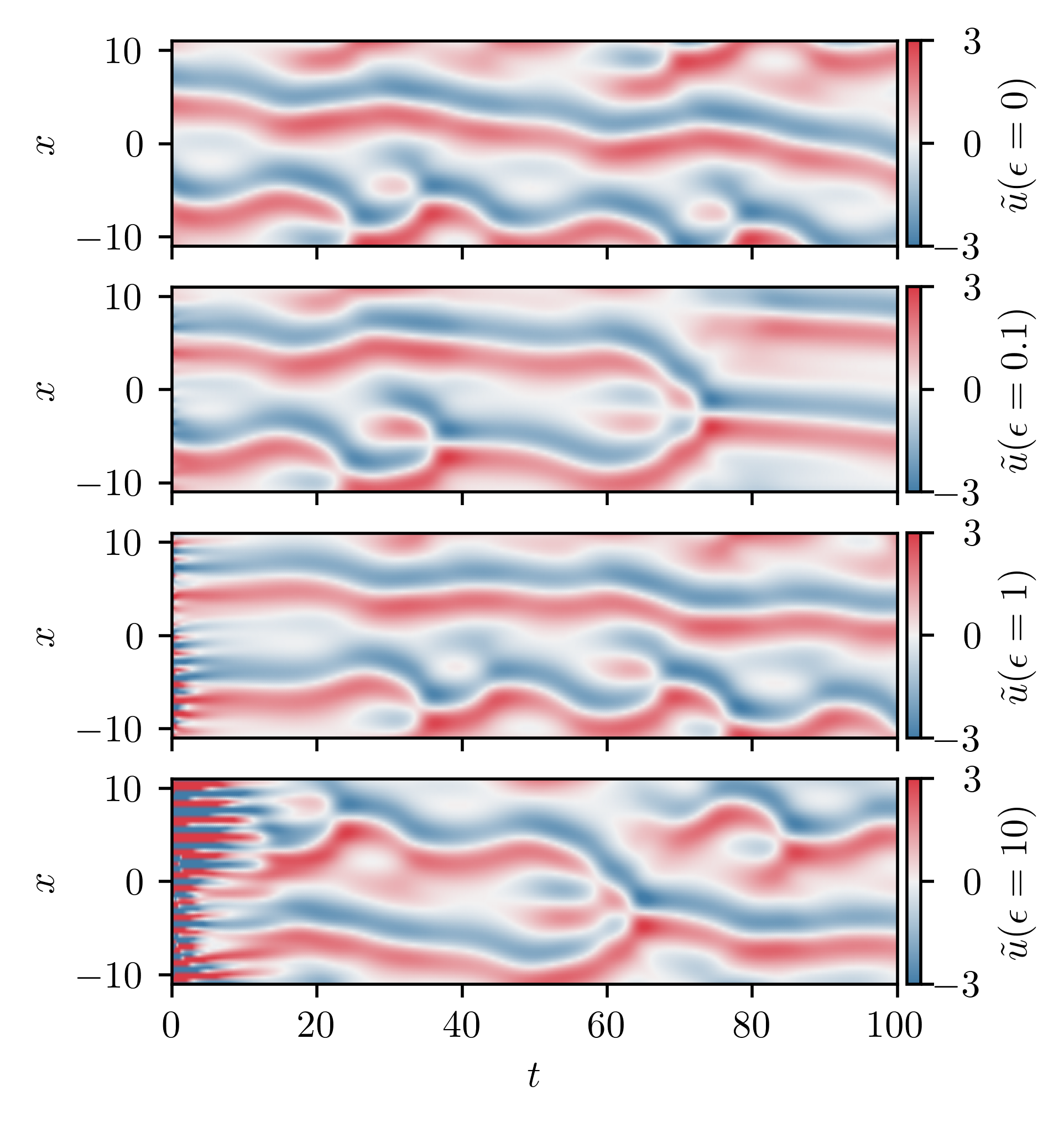}
		\begin{picture}(0,0)
		\put(-115,263){\contour{white}{ \textcolor{black}{a)}}}
		\put(-115,206){\contour{white}{ \textcolor{black}{b)}}}
		\put(-115,149){\contour{white}{ \textcolor{black}{c)}}}
		\put(-115,92){\contour{white}{ \textcolor{black}{d)}}}
		\end{picture}
		\caption{}
		\vspace{-10mm}
		\label{fig:Noisea}
	\end{subfigure}
	\begin{subfigure}[b]{0\textwidth}\caption{}\vspace{-10mm}\label{fig:Noiseb}\end{subfigure}\begin{subfigure}[b]{0\textwidth}\caption{}\vspace{-10mm}\label{fig:Noisec}\end{subfigure}\begin{subfigure}[b]{0\textwidth}\caption{}\vspace{-10mm}\label{fig:Noised}\end{subfigure}    
	\vspace{-1.5\baselineskip}
	\captionsetup{justification=raggedright}
	\caption{(a)-(d) are trajectories of the CNN model for initial conditions with $\epsilon=0$, $0.1$, $1$, and $10$, respectively.} 
	\label{fig:KSE_Noise} 
	\vspace{-5mm}
\end{figure}
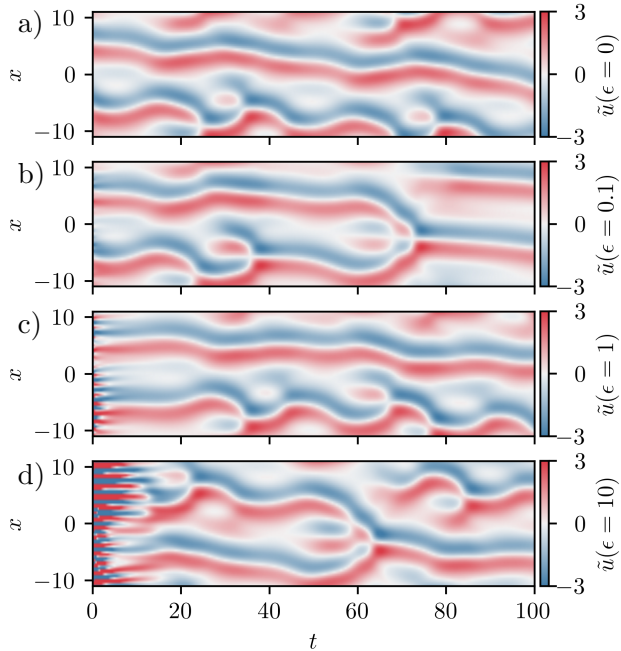

\ALrevise{Similar to the VBE, the} presence of a linear term \ALrevise{makes the KSE models highly robust} to noisy initial conditions. Figure \ref{fig:KSE_Noise} shows the short-time evolution of predictions from the CNN approach with varying levels of noise. Here the initial condition is perturbed with Gaussian noise of magnitude $\mathcal{N}(0,\epsilon^2)$ added to Fourier modes of $20^{\text{th}}$ up to $31^{\text{st}}$. Even with the extreme levels of noise shown in \ref{fig:Noised} the solution quickly decays back to the attractor. Despite the magnitude of the \ALrevise{CNN linear term being much smaller than the true linear term}, the \ALrevise{CNN} still rapidly damps high-wavenumbers.


The other useful aspect of \ALrevise{the stabilized} neural ODE architecture is \ALrevise{it lends itself naturally to reduced-order modeling (ROM).} 
\ALrevise{As mentioned in section \ref{sec:Framework}, a standard approach to ROM involves projecting the true ODE (Eq.\ \ref{eq:ODEFull}) onto a basis given by the leading $d_p$ eigenvectors of $A$. This is the resolved dynamics in Eq.\ \ref{eq:resolved}. Then, the unresolved dynamics, Eq.\ \ref{eq:unresolved} can be approximated by setting $q=0$, which is the standard Galerkin approach, or by 
using Eq.\ \ref{eq:q} to solve for $q$, which is the nonlinear Galerkin approach. Here we show we can successfully apply both these ROM methods to our CNN neural ODE model that we find strictly from data, whereas typically Eq.\ \ref{eq:ODEFull} must be known. We first compare the projection operator of the CNN neural ODE to the true projection operator, and then we compare the long-time statistical performance of the ROM when varying dimension $d_p$.}

\ALrevise{For these ROM approaches the eigenvector basis is typically selected from largest to smallest eigenvalues}. However, selecting the eigenvectors in this fashion does not necessarily \ALrevise{result in the best agreement with the underlying assumption of this ROM -- that} $\mathrm{d}q/\mathrm{d}t=0$. \ALrevise{To investigate this assumption, we consider the dynamics from projecting the equation onto each eigenvector $d\mathrm{v}_i^Tu/dt=\dot{\mathrm{v}}_i$}.
\ALrevise{The ROM assumption is reasonable when, for some $p$, $\dot{\mathrm{v}}_i\approx 0$ for all $i=p+1,\dots,d$.}

\begin{figure*}
	\centering
	\captionsetup[subfigure]{labelformat=empty}
	\begin{subfigure}[b]{17.2 cm}
		\includegraphics[trim=0 0 0 0,width=\textwidth,clip]{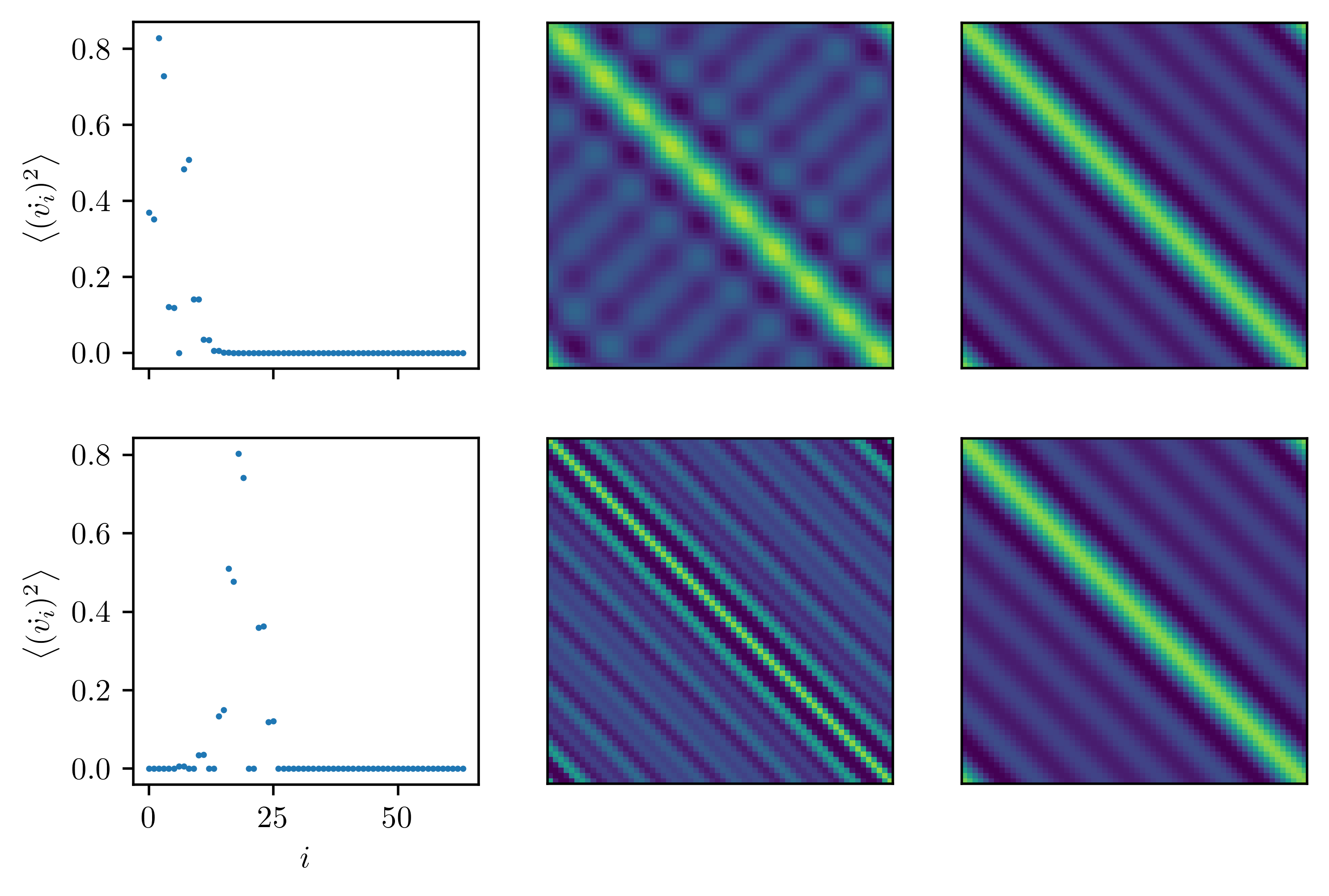}
		\begin{picture}(0,0)
		\put(-240,330){a)}
		\put(-60,330){b)}
		\put(93,330){c)}
		\put(-240,177){d)}
		\put(-60,177){e)}
		\put(93,177){f)}
		\end{picture}
		\caption{}
		\vspace{-10mm}
		\label{fig:ROM11}
	\end{subfigure}
	\begin{subfigure}[b]{0\textwidth}\caption{}\vspace{-10mm}\label{fig:ROM13}\end{subfigure}
	\begin{subfigure}[b]{0\textwidth}\caption{}\vspace{-10mm}\label{fig:ROM15}\end{subfigure}
	\begin{subfigure}[b]{0\textwidth}\caption{}\vspace{-10mm}\label{fig:ROM21}\end{subfigure}
	\begin{subfigure}[b]{0\textwidth}\caption{}\vspace{-10mm}\label{fig:ROM23}\end{subfigure}
	\begin{subfigure}[b]{0\textwidth}\caption{}\vspace{-10mm}\label{fig:ROM25}\end{subfigure}
	\captionsetup{justification=raggedright}
	\caption{(a)-(c) are the variance of the ODE projected onto each eigenvector, the unsorted projection operator, and the sorted projection operator, for the true linear term. (d)-(f) are the same plots for the linear term learned by the CNN. These projection operators are for 10 eigenvectors.}
	\label{fig:Proj}
	\vspace{-5mm}
\end{figure*}
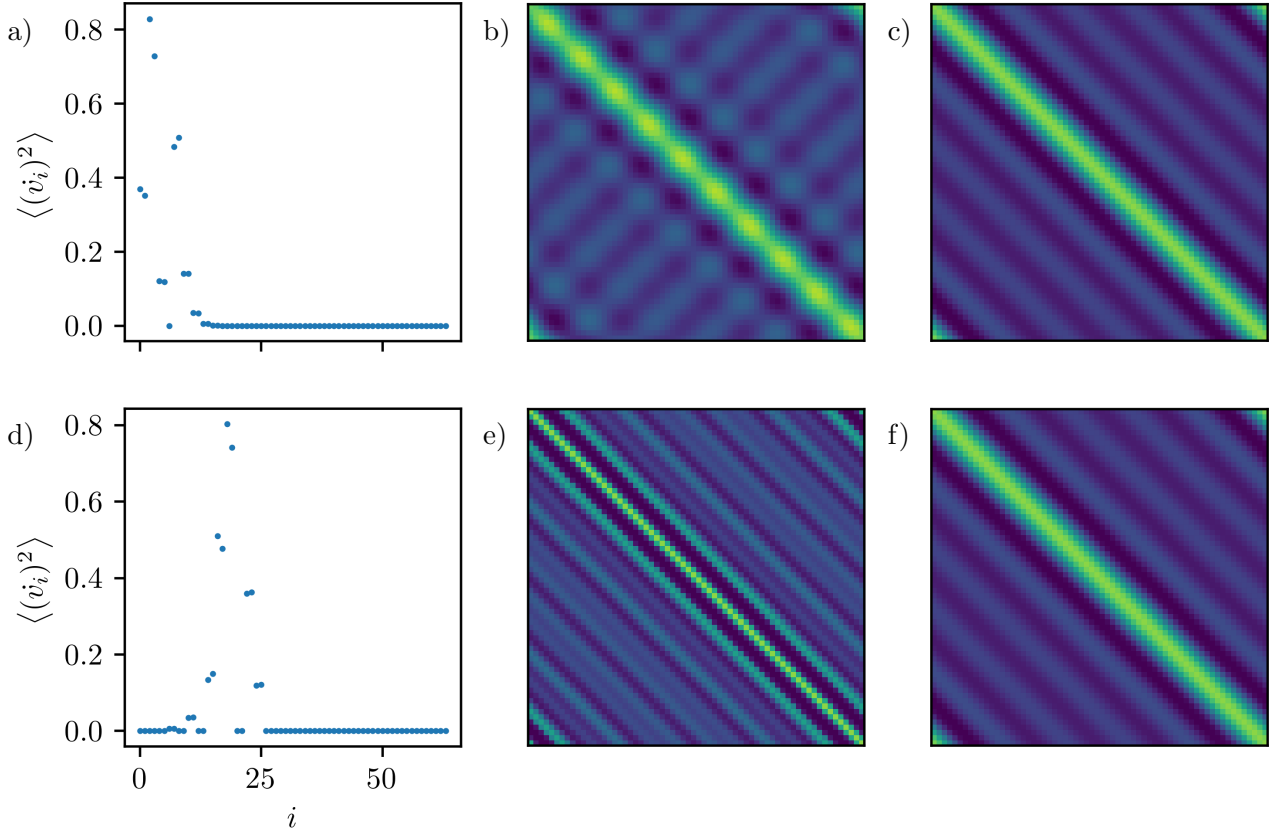

In Fig.\ \ref{fig:ROM11} we show the variance \ALrevise{in $\dot{\mathrm{v}}_i$} sorted \ALrevise{from largest to smallest eigenvalue $\lambda_i$.} \ALrevise{Each eigenvector can be considered separately because the linear term is symmetric in both the true system and from the CNN approach resulting in an orthogonal basis.}
Figure\ \ref{fig:ROM11} \ALrevise{shows sorting by eigenvalue} does not \ALrevise{lead to a monotonic decrease} in $\dot{\mathrm{v}}_i$. In Fig.\ \ref{fig:ROM21} we \ALrevise{apply this same sorting with the} linear term learned by the CNN\ALrevise{. Here the sorting is worse, with the leading terms showing little variance}. 
\ALrevise{In order for the CNN ROM to perform well the projection matrix must match the projection matrix for the true linear term.} 
In Fig.\ \ref{fig:ROM13} \ALrevise{we show a projection matrix for the true linear term and in Fig.\ \ref{fig:ROM23} we show a projection matrix for the CNN approach. Both of which were generated using the eigenvalue sorting and the first 10 eigenvectors.} These projection matrices are substantially different\ALrevise{, and the CNN projection matrix, in particular, is not useful for ROM.}

\ALrevise{In general, when only data is available there is no immediate way to compute $\dot{\mathrm{v}}_i$, however, because we learn the ODE directly from data computing $\dot{\mathrm{v}}_i$ after the fact is trivial. This motivates another reason why our stabilized neural ODE approach naturally lends itself to ROM. Instead of organizing according to eigenvalues, we can, instead, sort according to the variance of $\dot{\mathrm{v}}_i$, which more accurately reflects the assumption taken in nonlinear Galerkin. In Fig.\ \ref{fig:ROM15} we show the projection matrix from the true linear term with this sorting and in Fig.\ \ref{fig:ROM25} we show the projection matrix from the CNN linear term. The two projections matrices match nearly exactly.}

\begin{figure} 
    \centering
	\includegraphics[trim=0 0 0 0,width=8.6 cm,clip]{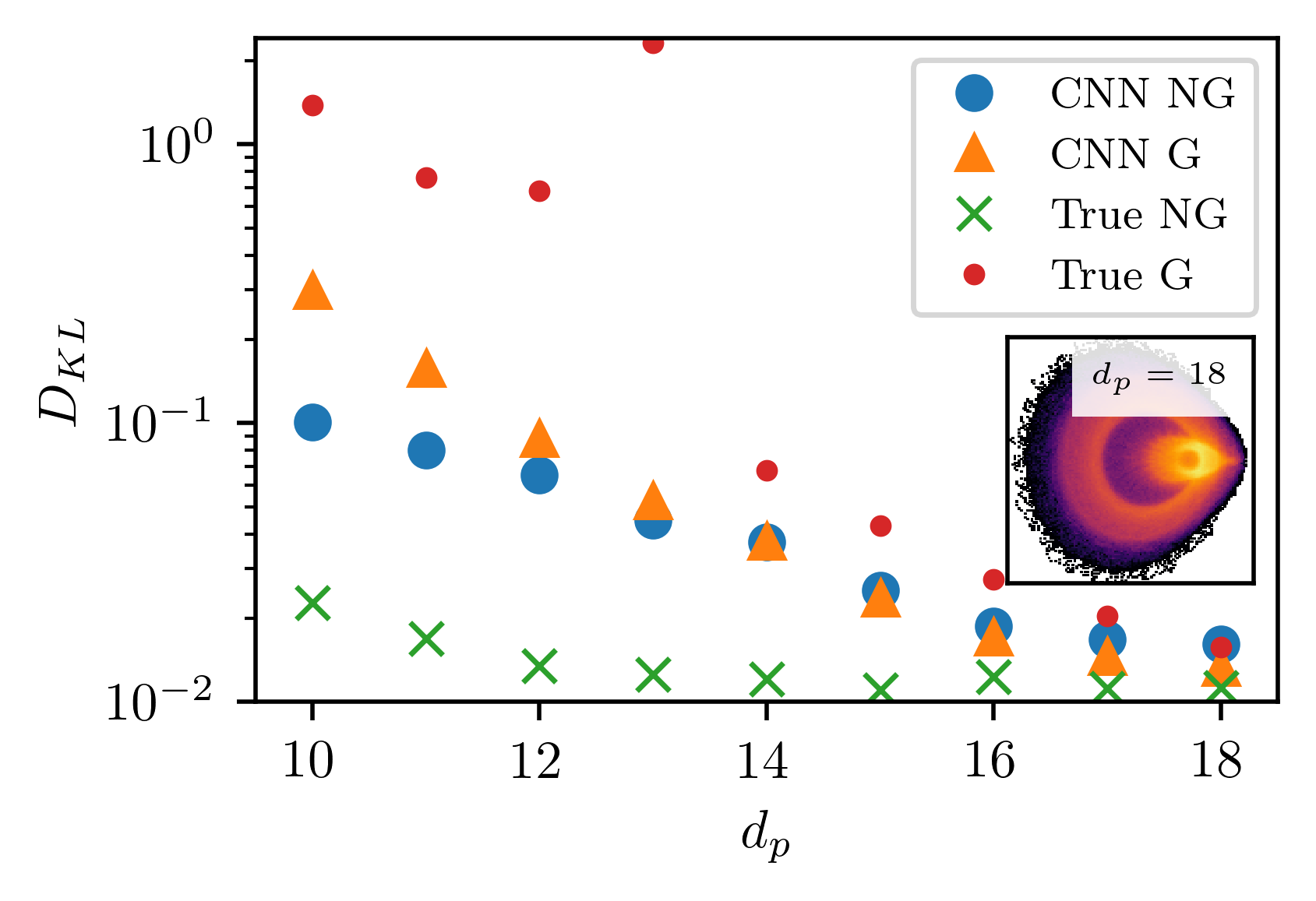}
	\captionsetup{justification=raggedright}
	\caption{KL divergence of the joint PDFs generated with either standard Galerkin ``G" or nonlinear Galerkin ``NG" when using the CNN neural ODE ``CNN" or the true equations ``True". The inset is the CNN Galerkin ROM at a dimension $d_p=18$.}
	\label{fig:KL}
\end{figure}

With the \ALrevise{projection} selected according to this variance criterion, we show the \ALrevise{standard Galerkin and nonlinear Galerkin approaches performance} in Fig.\ \ref{fig:KL}. \ALrevise{For this comparison we consider both the CNN neural ODE, sorted according to the variance criterion, and the true ODE, sorted by the standard eigenvalue criterion.} \ALrevise{We compare ROM performance} with the Kullback-Leibler (KL) divergence of the $u_x$ vs $u_{xx}$ joint PDF
\begin{equation}
	D_{KL}(\tilde{P}||P)=\int_{-\infty}^\infty \int_{-\infty}^\infty \tilde{P}(u_x,u_{xx}) \text{ln}\dfrac{\tilde{P}(u_x,u_{xx}) }{P(u_x,u_{xx})}\, du_x du_{xx}.
\end{equation}
In this equation $P$ is the true joint PDF and $\tilde{P}$ is the ROM joint PDF. When $P=0$ or $\tilde{P}=0$ we adopt the convention of setting the term in the integral to 0 as was done in \cite{Cazais2015}.

\ALrevise{In Fig.\ \ref{fig:KL} we vary the dimension $d_p$ (i.e. the number of modes on which we project) and compute the KL divergence between the true joint PDF and the ROM joint PDFs, generated the same way as in Fig.\ \ref{fig:KSE_PDF}.} \ALrevise{In the case of the Galerkin ROM, we see that the CNN approach does a better job than even the true Galerkin approach at reconstructing the joint PDF. In the case of nonlinear Galerkin, the CNN approach shows more error than the true nonlinear Galerkin approach, but performs better than the standard Galerkin approaches. For all ROMs we see excellent agreement around a dimension of $d_p\sim$18, as can be seen in the inset, above which there is little change in the KL divergence and the ROM agree well with the full system. These results show that by using classical ROM approaches on our data-driven stabilized neural ODEs we can achieve excellent model performance at far fewer dimensions than the underlying system.}


\section{Conclusion} \label{sec:Conclusion}

\ALrevise{Stabilizing} neural ODEs with a linear and a nonlinear term is an effective way to improve the predictive capabilities of these models. By implementing this structure we find the models are more robust to noise, perform better in capturing statistics, and provide a method for selecting a natural basis, different from conventional approaches, for reduced-order models. We test this setup first on the VBE, where the addition of the linear term results in a model that effectively captures the shock dynamics and the energy spectrum, along with generalizing to noisy initial conditions. We then test it on the KSE, where we find the linear term is necessary to keep trajectories on the attractor, and again \ALrevise{the linear term provides} excellent robustness to noise. With the KSE we further show \ALrevise{our stabilized neural ODE method lends itself naturally to ROM, and}  that, with proper sorting of the eigenvectors of the linear term, we can accurately recreate the joint PDF of $u_x$ and $u_{xx}$ with \ALrevise{far fewer dimensions than required for the full simulation.}

\ALrevise{The future directions of this research include apply the method to more complex problems, where the existence of an inertial manifold may be unknown, and investigating the importance of adding explicity constraints on the NNs for stability. In particular, we are immediately interested in applying this method to canonical fluid flow problems (e.g. Couette flow). This modeling could either be directly data-driven, as in this paper, or for closure of low resolution physics-based models. Regardless, with problems of increasing complexity it may require the use of adding additional constraints for stability. This can be done by enforcing that the linear term has a specific eigenvalue spectrum, or by tuning the Lipschitz constant of the nonlinearity. For example, we could enforce the spectral gap condition, which would guarantee the existence of an inertial manifold.}


\begin{acknowledgments}
This work was supported by the U.S. Department of Energy (DOE), Office of Science, Office of Advanced Scientific Computing Research (ASCR), under Contract No. DE-AC02–06CH11357, at Argonne National Laboratory, and by the Office of Fusion Energy Sciences and ASCR under the Scientific Discovery through Advanced Computing (SciDAC) project of Tokamak Disruption Simulation at Los Alamos National Laboratory. We acknowledge funding support from ASCR for DOE-FOA-2493 ``Data-intensive scientific machine learning". This research used resources of the Argonne Leadership Computing Facility, which is a DOE Office of Science User Facility supported under Contract No. DE-AC02–06CH11357.
\end{acknowledgments}

\bibliography{library.bib}

\end{document}